\documentclass[manuscript]{acmart}

\usepackage{latexsym}
\usepackage[T1]{fontenc}
\usepackage[utf8]{inputenc}
\usepackage{microtype}
\usepackage{enumitem}

\usepackage{hyperref}
\usepackage{url}
\usepackage{booktabs}
\usepackage{amsfonts}
\usepackage{nicefrac}
\usepackage{microtype}
\usepackage{xcolor}

\usepackage{graphicx}
\usepackage{tikz}
\usepackage{pgfplots}
\usepackage{pgfplotstable}
\usetikzlibrary{patterns}
\usepgfplotslibrary{colormaps}
\pgfplotsset{compat=1.18}
\usetikzlibrary{pgfplots.colormaps}
\usepackage{pgf-pie} 
\usepackage{xcolor,colortbl}
\usepackage{multirow}
\usepackage{xurl}
\usepackage{siunitx}
\usepackage{enumitem}
\usepackage{dcolumn}
\usepackage{makecell}

\usetikzlibrary{pgfplots.statistics}
\usetikzlibrary{matrix}
\usetikzlibrary{calc}

\newcolumntype{d}[1]{D{.}{.}{#1}}

\newcommand{\gray}[1]{%
{\cellcolor{gray!\number\numexpr7*#1/12\relax}}%
}

\setitemize{noitemsep}

\AtBeginDocument{%
  \providecommand\BibTeX{{%
    \normalfont B\kern-0.5em{\scshape i\kern-0.25em b}\kern-0.8em\TeX}}}

\acmYear{2024}\copyrightyear{2024}
\setcopyright{rightsretained}
\acmConference[ACM FAccT '24]{ACM Conference on Fairness, Accountability, and Transparency}{June 3--6, 2024}{Rio de Janeiro, Brazil}
\acmBooktitle{ACM Conference on Fairness, Accountability, and Transparency (ACM FAccT '24), June 3--6, 2024, Rio de Janeiro, Brazil}
\acmDOI{10.1145/3630106.3658966}
\acmISBN{979-8-4007-0450-5/24/06}
\acmPrice{}

\begin{document}

\title{Laboratory-Scale AI: Open-Weight Models are Competitive with ChatGPT Even in Low-Resource Settings}

\author{Robert Wolfe}
\affiliation{
  \institution{University of Washington}
  \city{Seattle}
  \state{Washington}
  \country{United States}}
\email{rwolfe3@uw.edu}

\author{Isaac Slaughter}
\affiliation{
  \institution{University of Washington}
  \city{Seattle}
  \state{Washington}
  \country{United States}}
\email{is28@uw.edu}

\author{Bin Han}
\affiliation{
  \institution{University of Washington}
  \city{Seattle}
  \state{Washington}
  \country{United States}}
\email{bh193@uw.edu}

\author{Bingbing Wen}
\affiliation{
  \institution{University of Washington}
  \city{Seattle}
  \state{Washington}
  \country{United States}}
\email{bingbw@uw.edu}

\author{Yiwei Yang}
\affiliation{
  \institution{University of Washington}
  \city{Seattle}
  \state{Washington}
  \country{United States}}
\email{yanyiwei@uw.edu}

\author{Lucas Rosenblatt}
\affiliation{
  \institution{New York University}
  \city{New York}
  \state{New York}
  \country{United States}}
\email{lucas.rosenblatt@nyu.edu}

\author{Bernease Herman}
\affiliation{
  \institution{University of Washington}
  \city{Seattle}
  \state{Washington}
  \country{United States}}
\email{bernease@uw.edu}

\author{Eva Brown}
\affiliation{
  \institution{University of Washington}
  \city{Seattle}
  \state{Washington}
  \country{United States}}
\email{evamxb@uw.edu}

\author{Zening Qu}
\affiliation{
  \institution{University of Washington}
  \city{Seattle}
  \state{Washington}
  \country{United States}}
\email{zqu@uw.edu}

\author{Nic Weber}
\affiliation{
  \institution{University of Washington}
  \city{Seattle}
  \state{Washington}
  \country{United States}}
\email{nmweber@uw.edu}

\author{Bill Howe}
\affiliation{
  \institution{University of Washington}
  \city{Seattle}
  \state{Washington}
  \country{United States}}
\email{billhowe@uw.edu}

\renewcommand{\shortauthors}{Wolfe et al.}
\renewcommand{\shorttitle}{Laboratory-Scale AI}

\begin{abstract}
  The rapid proliferation of generative AI has raised questions about the competitiveness of lower-parameter, locally tunable, open-weight models relative to high-parameter, API-guarded, closed-weight models in terms of performance, domain adaptation, cost, and generalization. Centering under-resourced yet risk-intolerant settings in government, research, and healthcare, we see for-profit closed-weight models as incompatible with requirements for transparency, privacy, adaptability, and standards of evidence.  Yet the performance penalty in using open-weight models, especially in low-data and low-resource settings, is unclear. 
  
  We assess the feasibility of using smaller, open-weight models to replace GPT-4-Turbo in zero-shot, few-shot, and fine-tuned regimes, assuming access to only a single, low-cost GPU.  We assess value-sensitive issues around bias, privacy, and abstention on three additional tasks relevant to those topics. We find that with relatively low effort, very low absolute monetary cost, and relatively little data for fine-tuning, small open-weight models can achieve competitive performance in domain-adapted tasks without sacrificing generality.  We then run experiments considering practical issues in bias, privacy, and hallucination risk, finding that open models offer several benefits over closed models. We intend this work as a case study in understanding the opportunity cost of reproducibility and transparency over for-profit state-of-the-art zero shot performance, finding this cost to be marginal under realistic settings.

\end{abstract}



\maketitle

\section{Introduction}

\begin{figure*}[h]
    \includegraphics[width=\textwidth]{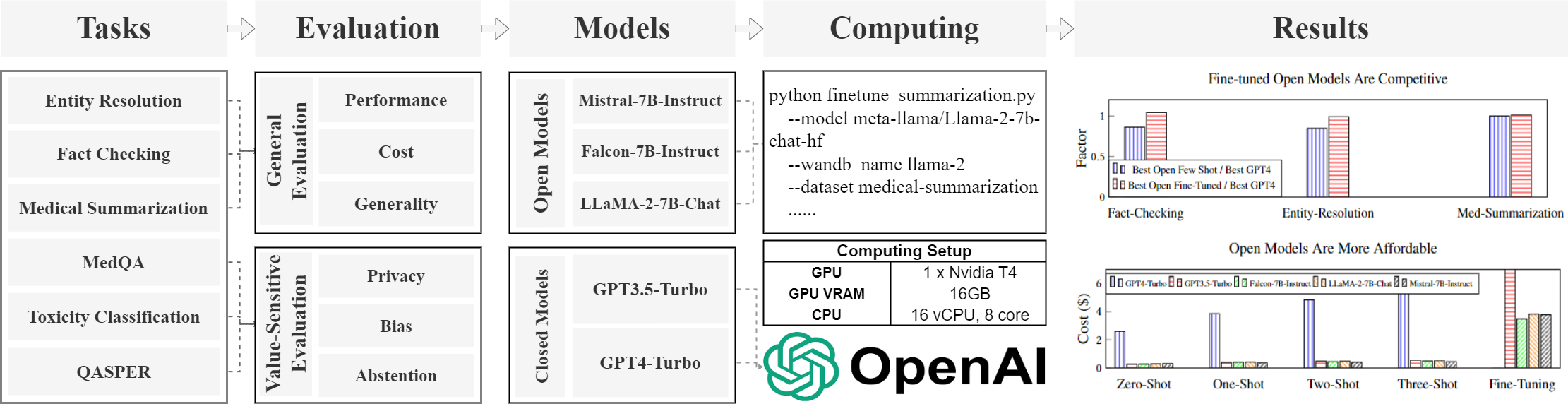}
    \caption{\footnotesize We compared domain-specific performance, general-purpose usability, and amenability to responsible use of three open language models with two dominant closed models. We found that fine-tuning open models renders them competitive with few-shot closed models at low cost.\looseness=-1}
\end{figure*}

According to a November 2023 report by The Verge, OpenAI’s ChatGPT boasts more than one hundred million weekly users, two million developers using the API, and more than 80\% adoption of among Fortune 500 companies, making it one of fastest growing services in history~\cite{porter2023chatgpt}. Despite the influence of OpenAI's flagship language models on the world’s ways of working and seeking information, scientists know little about them: details of the architecture, parameter counts, and training data of GPT-3.5-Turbo and GPT-4-Turbo are omitted or glancingly described in the company's technical reports \cite{openai2023gpt4}. Reaffirming the ``values encoded in machine learning research'' described by \citet{birhane2022values}, transparency has taken a back seat to values that preserve corporate competitive advantage. For many scientists and public interest practitioners, this lack of transparency is at best concerning, and often a reason to avoid such models in their work altogether \cite{liesenfeld2023opening,palmer2023using}. At the same time, recent research has enabled the use of accessible and inexpensive hardware to train domain-adapted models. Eight-bit and four-bit quantization allow very large models to run on affordable commercial-grade GPUs \cite{dettmers2022llm, dettmers2023case}. Quantized low-rank adaptation (qLoRA) \cite{hu2021lora,dettmers2023qlora} allow large models to be customized to a domain by adding and tuning a modest number of parameters while allowing most pretrained weights to remain fixed. \looseness=-1

These technologies could collectively help enable a future for AI that is not wed to the interests of Big Tech corporations --- one that prioritizes transparency, cost-efficiency, and the domain-specific and responsible application of language technologies, in addition to strong performance. In this work, we intend to provide an empirical, practical foundation for this approach, which we call "Laboratory-Scale AI." Concretely, we address the following research questions:

\noindent \textit{\textbf{RQ1}: Do open models offer domain-specific performance competitive with closed models for tasks of scientific and public interest?} We assess open models against closed models on three tasks selected for their scientific or public interest value: government records entity resolution \cite{10.14778/2367502.2367564}, climate misinformation fact-checking \cite{diggelmann2020climate}, and clinical dialogue summarization \cite{mts-dialog}. We evaluate OpenAI’s GPT-3.5-Turbo and GPT-4-Turbo against three open instruction-tuned models: Mistral-7b-Instruct-v.01 \cite{jiang2023mistral}, Falcon-7b-Instruct \cite{almazrouei2023falcon}, and LLaMA-2-Chat-7b \cite{touvron2023llama}. Results show GPT-4-Turbo exceeds the performance of the four other models when using them in a few-shot setting, but GPT-3.5-Turbo and \textbf{open models are comparable to GPT-4-Turbo or exceed its performance after fine-tuning for a single dataset epoch}. On fact-checking, fine-tuned Mistral-7b-Instruct achieves accuracy of .75, exceeding the mark of .72 by three-shot GPT-4-Turbo. \looseness=-1

\noindent \textit{\textbf{RQ2}: Are open models cost-competitive with closed models?} We find that the cost of running inference on a test dataset with GPT-4-Turbo is comparable to both fine-tuning and inference using an open model. Cost savings achieved by using an open model after fine-tuning are especially notable: on the climate fact-checking test dataset, \textbf{inference is almost ten times less costly using an open model} (Mistral-7B-Instruct, \$.31) than zero-shot GPT-4-Turbo (\$2.65). \looseness=-1

\noindent \textit{\textbf{RQ3}: How responsive are small open models to domain-specific fine-tuning data?} We evaluate the performance of LLaMA-2-Chat-7B fine-tuned for clinical dialogue summarization after 0\%, 20\%, 40\%, 60\%, 80\% and 100\% of task training data, and evaluate the performance of the LLaMA-2-Chat-7B, Falcon-7B-Instruct, and Mistral-7B-Instruct every 500 steps of the 4,298-sample fact-checking dataset. After 20\% of the fine-tuning dataset (240 samples), the summarization model achieves accuracy of .79 on the test dataset, within .02 of its best. After 2,000 fact-checking samples, Mistral-7B-Instruct achieves accuracy of .71, comparable to fine-tuned GPT-3.5-Turbo. The results show \textbf{open models can be adapted with a small amount of data}, without the need for large-scale data collection. \looseness=-1

\noindent \textit{\textbf{RQ4}: Can fine-tuned, domain-specific models provide a general-purpose chat-based interface for end users?} Chat-based language models provide an approachable interface to end users. We investigate whether fine-tuning inhibits the utility of this interface by comparing the performance of fine-tuned LLaMA-2-Chat-7B models with the base model on tasks not reflected in the model's fine-tuning dataset. For example, we measure performance of the fine-tuned fact-checking model on the entity resolution task. We find that \textbf{fine-tuned open models exhibit performance comparable to general-purpose base chat models}, and in some cases exceed it: for example, the fact-checking LLaMA-2 improves to .85 accuracy on entity resolution, from the base model's .77. \looseness=-1

\noindent \textit{\textbf{RQ5}: Can laboratory-scale language models be used in a responsible manner?} We evaluate open and closed models on three tasks of  importance for the ethical application of instruction-tuned language models: question answering under differentially private fine-tuning, demographic bias in toxic comment classification, and abstention from answering questions for which insufficient information is available to answer correctly. We find that the performance of open models fine-tuned using a private optimizer approaches non-private fine-tuning, suggesting a better privacy alternative to closed models; that open models exhibit moderate bias that fine-tuning largely fails to mitigate; and that fine-tuning open models can improve their abstention properties: fine-tuned LLaMA-2-7B-Chat achieves an abstention score of .99 (maximum 1.0), exceeding the performance of fine-tuned GPT-3.5-Turbo. \textbf{While open models exhibit greater bias, they offer greater privacy affordances than closed models, and in some cases abstain more reliably after fine-tuning.} \looseness=-1

Our experiments demonstrate that a fine-tuned open model running on inexpensive hardware can exceed the performance of GPT-4-Turbo at lower cost. In addition to our core empirical contributions, we offer a practical discussion of the challenges and opportunities of adopting a laboratory-scale approach in the Discussion section.

\section{Related Work}
We review the related work on generative language models, with attention to model accessibility and evaluation.

\noindent \textbf{Generative Language Models} Generative Pretrained Transformers (GPTs) adopt a modified transformer deep learning architecture \cite{vaswani2017attention}, employing decoder layers to generate an output conditioned on the preceding input \cite{radford2018improving}. GPT language models \cite{radford2019language, brown2020language, openai2022chatgpt} are pretrained on the ``causal'' language modeling objective, taking as input a series of subword tokens and generating the predicted next token \cite{dai2015semi}. As a result, GPTs can generate freeform output, responsive to a user's input \cite{radford2018improving}. Early applications of GPTs demonstrated few-shot prompting capabilities: given examples of desired behavior at inference (rather than in training), the model generates a task-relevant output \cite{brown2020language}. More sophisticated prompting strategies like chain-of-thought also provide explanations of the reasoning process to derive correct answers \cite{wei2022chain}. A ``zero-shot'' setting refers to asking the model to produce correct output with no examples \cite{wang2019survey}.

While remarkably effective for NLP tasks, models that require prompts with examples in context generally serve as a poor interface for lay users, and can result in unpredictable model behavior \cite{dang2022prompt}. To provide a more naturalistic and reliable interface, \citet{ouyang2022training} trained GPT models to adhere to the natural language instructions of a human user, with or without examples. Models like OpenAI's ChatGPT \cite{openai2022chatgpt} and Meta's LLaMA \cite{touvron2023llama} are fine-tuned to engage in dialogue, producing a “chat” based interface wherein the model and a human user take turns, with the user typically providing an instruction or request \cite{ray2023chatgpt}. Such models are typically trained to be safe and helpful to the end user via reinforcement learning from human feedback (RLHF) \cite{bai2022training}, or direct optimization using a language model as the reward model \cite{rafailov2023direct}. Our research studies decoder-only generative language models fine-tuned to follow instructions of human users. \looseness=-1

\noindent \textbf{Improving Model Accessibility}
Most generative language models use billions of trainable parameters to mimic human language \cite{touvron2023llama2,jiang2023mistral,almazrouei2023falcon}. Training or deploying such models requires vast financial resources, making them difficult to train and access for researchers and public interest practitioners \cite{bender2021dangers}. However, while pretraining LLMs remains prohibitively expensive, recent techniques mitigate the difficulties of using models with low-cost hardware. For example, quantization loads a model in a lower level of precision than used during pretraining \cite{dettmers2022llm}. While most generative models are pretrained in 32-bit or mixed 32/16-bit precision, quantization loads the weights in 8-bit \cite{dettmers2022llm}, 4-bit \cite{dettmers2023case}, or even 2-bit precision \cite{dettmers2023case, chee2023quip}, reducing memory demands. Because transformer models achieve much greater speed using GPUs, the bottleneck for efficiently deploying a model is often the amount of memory (Video-RAM) on the GPU device \cite{dettmers2023qlora}. However, quantization alone does not permit efficient training on commercial grade hardware \cite{dettmers2022llm}. Methods known as \textit{parameter efficient fine-tuning} (PEFT) attempt to preserve the general-purpose functionality of a pretrained model while adapting it for a specific task \cite{zaken2022bitfit, ding2023parameter}. Among the most widely used PEFT techniques is low-rank adaptation (LoRA) \cite{hu2021lora}. LoRA inserts small, trainable weight matrices into a pretrained model, which are fine-tuned while leaving the learned parameters of the pretrained model unchanged \cite{hu2021lora}, reducing fine-tuning memory costs. LoRA weights also require less space than a fully fine-tuned model \cite{hu2021lora}. Saving a fine-tuned LLaMA-2-7B-Chat model would require about 13.5GB of storage; saving only the LoRA weights --- which can later be inserted into the pretrained model --- requires only around 260MB \cite{hu2021lora, dettmers2023qlora}. \citet{dettmers2023qlora} introduced qLoRA, a method for allowing trainable LoRA weights to be inserted into quantized models, allowing relatively large models to be mounted on a small GPU and customized using LoRA \cite{dettmers2023qlora}. \looseness=-1

\noindent \textbf{Language Model Benchmarking and Evaluation}
Language models are typically evaluated on specific tasks intended to measure performance on a function of interest, such as sentiment analysis or machine translation. Benchmarks for language models consist of collections of these tasks that assess model ability in a wider domain  \cite{colombo2022best}. For example, the Massive Multitask Language Understanding (MMLU) benchmark measures scientific and world knowledge acquired during pretraining \cite{hendrycks2021measuring}. At a higher level, Stanford's Holistic Evaluation of Language Models (HELM) collects tasks and benchmarks, and asks human users to competitively evaluate model responses to user inputs \cite{liang2023holistic}. Efforts to use human preferences to evaluate models in the wild include LMSYS Chatbot Arena, where users interact with two anonymous language models and vote for a preferred model \cite{zheng2023judging}. \citet{bommasani2023foundation} introduce the Foundation Models Transparency Index, which scores models on transparency-related metrics such as model availability and training details. \looseness=-1

\noindent \textbf{Benchmarking ChatGPT} The popularity of ChatGPT has prompted researchers to evalute it against traditional NLP methods and models. \citet{kocon2023chatgpt} evaluate ChatGPT against state-of-the-art NLP models on 25 NLP tasks, finding that GPT-3.5-Turbo and GPT-4 are outperformed by these models and methods. \citet{thalken2023modeling} show that a fine-tuned LEGAL-BERT \cite{chalkidis2020legal} is the best-performing model for classifying legal reasoning, outperforming models like GPT-4 and LLaMA-2-Chat. \citet{liang-etal-2023-breaking} find that fine-tuned sentence transformer models outperform few-shot GPT-3.5-Turbo and GPT-4 on a financial text classification task. \citet{wang2023chatgpt} find that a fine-tuned BERT can outperform ChatGPT on sentiment analysis. We build on prior work in benchmarking by comparing open and closed models, but differ by focusing on autonomy, transparency, and responsible use as much as performance.


\section{Approach}

We review the models studied, evaluations employed, and consistent cloud environment used across our experiments.

\subsection{Models}

The models studied share the following characteristics:
\begin{itemize}[leftmargin=4mm]
    \item \textbf{Causal (Generative) Pretraining Objective}: All models share the causal language modeling (next-word prediction) objective introduced to the transformer architecture by \citet{radford2018improving}. 
    \item \textbf{Instruction-Following}: All models undergo supervised fine-tuning to enable a user to issue instructions in natural language, and receive a natural language response from the model \cite{ouyang2022training}.
    \item \textbf{7-Billion Parameters (Open Models)}: The open models each have approximately seven billion trainable parameters, allowing them to be deployed on identical cloud instances. OpenAI has not disclosed parameter counts for GPT-3.5-Turbo and GPT-4-Turbo, but studies suggest they are much larger than open models \cite{gpt4architecture}.
\end{itemize}

\noindent We study only generative, instruction-following models for three reasons. First, this accords with the architecture and training regimen of the closed, industry-dominant OpenAI models against which we assess open models. Second, both the closed and open models studied are among the most widely used language models in the world as of this writing, with Meta's LLaMA-2 model and Mistral's Instruct model routinely among the \href{https://huggingface.co/models?sort=trending}{most popular models} in the HuggingFace Transformers Python library. Third, these models provide an approachable natural language interface for users who may not be skilled in machine learning but would nonetheless benefit from the use of a domain-aligned language model. In addition to being aligned with our goal of empowering scientists and public interest users, the importance of an accessible interface is borne out by the success of ChatGPT, which far exceeds the userbase of OpenAI's own GPT-3 base models \cite{openaistats}. Finally, studying one group of similar models permits use of consistent infrastructure, allowing us to evaluate cost. \looseness=-1

\subsubsection{Defining Closed vs. Open Models}\label{sec:closeddef}

We define a \textit{closed} model as a model which is accessible only via a call to an API, and the weights and architecture of the model cannot be accessed. An \textit{open} model is one for which the pretrained weights and architecture are made available and can be modified and built upon. These models are not necessarily licensed to permit any use of the model, as such licenses may still prohibit commercialization or use for unethical purposes as defined by the organization releasing the weights \cite{touvron2023llama,touvron2023llama2}; that is, \textit{open} models are not necessarily fully \textit{open source} models. This definition of ``open'' aligns with that employed by \citet{palmer2023using} and \citet{rogers-etal-2023-closed}, but omits the requirement that researchers know the data on which the open model was trained, as even in previous definitions, data requirements come with the caveat that such data need not actually be ``available for direct inspection'' \cite{palmer2023using}. \looseness=-1

\subsubsection{Closed Models}

We study two closed OpenAI models: GPT-3.5-Turbo and GPT-4-Turbo.

\begin{itemize}[leftmargin=4mm]
    \item \textbf{OpenAI GPT-3.5-Turbo}: OpenAI's cost-efficient and broadly performant model optimized to follow instructions \cite{openai2022chatgpt,openaimodels}. A GPT-3.5-Turbo fine-tuned with RLHF and Proximal Policy Optimization is the model available to non-paying users who access ChatGPT through the online interface rather than the OpenAI API \cite{openai2022chatgpt}. We used OpenAI's default GPT-3.5-Turbo at the time of our experiments, which points to "gpt-3.5-turbo-0613" \cite{openaimodels}. \looseness=-1
    \item \textbf{OpenAI GPT-4-Turbo}: OpenAI's state-of-the-art language model, available at greater cost than GPT-3.5-Turbo \cite{openaimodels,openaipricing}. GPT-4-Turbo holds the \href{https://crfm.stanford.edu/helm/lite/latest/\#/leaderboard}{zero-shot state-of-the-art} on numerous NLP tasks as of this writing, and achieves first place in human evaluations of chat-based models in \href{https://huggingface.co/spaces/lmsys/chatbot-arena-leaderboard}{Chatbot Arena} \cite{zheng2023judging}. GPT-4-Turbo handles much longer text input sequences (128,000 tokens) than GPT-3.5-Turbo, as well as multiple input modalities, such as images \cite{openaimodels}.
\end{itemize}

\subsubsection{Open Models}

We study the following three open models:

\begin{itemize}[leftmargin=4mm]
    \item \textbf{TII Falcon-7B-Instruct}: A generative model pretrained on 1.5 trillion tokens of the RefinedWeb dataset \cite{penedo2023refinedweb}, released under the Apache 2.0 license by the UAE's Technology Innovation Institute (TII) in April 2023 \cite{almazrouei2023falcon}. TII's RefinedWeb dataset consists of filtered web data, and a subset is publicly available \cite{penedo2023refinedweb}. \looseness=-1
    \item \textbf{Meta LLaMA-2-7B-Chat}: A generative model pretrained on two trillion tokens of publicly available datasets and made available under the LLaMA 2 Community License by Meta AI in July 2023 \cite{touvron2023llama2}. The Chat model was fine-tuned for dialogue and underwent RLHF to improve helpfulness and minimize toxic output \cite{touvron2023llama2}.
    \item \textbf{Mistral AI Mistral-7B-Instruct-v0.1}: A generative model released under the Apache 2.0 license by Mistral AI in September 2023 \cite{jiang2023mistral}. Mistral-7B-Instruct-v0.1 is trained on an undisclosed quantity of data from the open internet, and exceeds LLaMA-2-7B-Chat and LLaMA-2-13B-Chat on common benchmarks  \cite{jiang2023mistral}.
\end{itemize}

\subsection{Model Evaluation}
We evaluate models in zero-shot, few-shot, and fine-tuned settings.

\begin{itemize}[leftmargin=4mm]
    \item \textbf{Zero-Shot}: The model is provided with a bare instruction of the task, and given the data to perform the task.
    \item \textbf{Few-Shot}: The model is provided with an instruction and examples of how to respond. We use multi-turn formatting to provide few-shot examples to the model, following the HuggingFace \href{https://huggingface.co/docs/transformers/chat_templating}{chat template documentation} for open models, and \href{https://platform.openai.com/docs/guides/prompt-engineering/strategy-write-clear-instructions}{OpenAI's documentation} for closed models. Falcon-7b-Instruct is not fine-tuned with a defined chat template, and we adhere to the \href{https://huggingface.co/tiiuae/falcon-7b-instruct/discussions/1\#64708b0a3df93fddece002a4}{guidance of the model's developers}, including examples in a single user prompt.
    \item \textbf{Fine-Tuned}: The model is fine-tuned on a task-specific dataset before evaluation on the task's test dataset. For consistency, we fine-tune for a single dataset epoch, reporting total examples in train and test datasets. We are unable to fine-tune GPT-4-Turbo, for which fine-tuning is available only via an \href{https://platform.openai.com/docs/guides/fine-tuning}{experimental program}. \looseness=-1
\end{itemize}

\subsubsection{Hyperparameters}

We employ four-bit quantization \cite{dettmers2023qlora} in both inference and fine-tuning. We use qLoRA adapters \cite{hu2021lora, dettmers2023qlora}  to fine-tune on domain-specific data, adopting the optimal hyperparameters specified by \citet{dettmers2023qlora}. Specifically, we use qLoRA to tune linear layers, set LoRA matrix rank to 32, and set LoRA dropout to .05, which improves performance in models with fewer than 13-billion parameters \cite{dettmers2023qlora}. We used gradient checkpointing during fine-tuning to save memory by recomputing activations during the model's backward pass \cite{chen2016training,gradientcheckpointing}. We set batch size to 1 due to memory limitations. We use the default hyperparameters for training GPT-3.5-Turbo, with the exception of fine-tuning for only one dataset epoch, rather than the OpenAI default of three. 



\begin{table*}[htbp]
    \centering
    \small
    \begin{tabular}{|c|c|c|c|c|}
       \toprule
       Representative Task & Train Samples & Validation Samples & Test Samples & Eval Metrics \\
       \midrule
       Entity Resolution & 700 & 100 & 200 & Accuracy, F1 Score\\
       Climate Fact Checking & 4,298 & 1,842 & 1,535 & Accuracy, Weighted F1 \\
       Clinical Dialogue Summarization & 1,201 & 100 & 400 & BLEU \cite{papineni2002bleu}, BERTScore F1 \cite{zhang2019bertscore} \\
       \bottomrule
    \end{tabular}
    \caption{\footnotesize Representative tasks with total training, validation, and test samples, as well as evaluation metrics.}
    \label{tab:performance_tasks}
\end{table*}

\subsection{Cloud Infrastructure}

We use a consistent cloud environment, allowing comparison of the cost and runtime of open vs. closed models. We defined the environment such that a 7-billion parameter model could be fine-tuned using qLoRA in 4-bit precision with a 1,024-token context window. We chose this setup because 7-billion parameter models are the lowest entry point for the three families of models we study (LLaMA-2-Chat-7B, Falcon-7B-Instruct, and Mistral-7B-Instruct), because fine-tuning in four-bit precision is competitive with fine-tuning in higher precision \cite{dettmers2023qlora}, and because our tasks (\textit{e.g.},  summarization), benefit from a context window of at least 1,000 tokens. Fine-tuning used a \$0.32 per hour Google Cloud Platform (GCP) \cite{bisong2019overview} spot instance with the following characteristics: a 16GB Nvidia T4 GPU; 60GB RAM; a 16vCPU, 8-core processor; and 200GB disk. While cost may vary based on region and provider, we found price was generally consistent on GCP and other providers such as AWS and Lambda Labs, within about \$.05 per hour. Because we expect that most laboratory-scale AI applications will be fault-tolerant during fine-tuning, we use spot instances, which may be terminated to support higher paying workloads, but are less costly than on-demand resources. \looseness=-1

\begin{table*}[htbp]
    \centering
    \footnotesize
    \tabcolsep=0.32cm
    \begin{tabular}{|l|l|c|c|c|c||c|c|}
    \hline
    \multirow{2}{*}{Model} & \multirow{2}{*}{Scenario} & \multicolumn{2}{|c|}{Entity-Resolution} & \multicolumn{2}{c||}{Fact-Checking} & \multicolumn{2}{c|}{Med-Summarization} \\ 
    \cline{3-8}
    & & Acc & F1 & Acc & F1 & BLEU & BERT-F1 \\
    \hline 
    \hline
        \multirow{4}{*}{GPT-4-Turbo} & Zero-Shot & \gray{88}0.93 & \gray{90}0.94 & \gray{95}0.72 & \gray{96}0.72 & \gray{55}0.06 & \gray{54}0.78 \\ 
    & One-Shot & \gray{88}0.93 & \gray{89}0.94 & \gray{95}0.72 & \gray{93}0.70 & \gray{78}0.08 & \gray{62}0.79 \\ 
    & Two-Shot & \gray{100}\textbf{0.97} & \gray{100}\textbf{0.98} & \gray{87}0.67 & \gray{90}0.68 & \gray{78}0.08 & \gray{65}0.80 \\ 
    & Three-Shot & \gray{98}\textbf{0.97} & \gray{99}0.97 & \gray{95}0.72 & \gray{96}0.72 & \gray{78}0.08 & \gray{66}0.80 \\ 
    \hline 
    \multirow{5}{*}{GPT-3.5-Turbo} & Zero-Shot & \gray{40}0.75 & \gray{49}0.78 & \gray{48}0.43 & \gray{49}0.42 & \gray{35}0.05 & \gray{39}0.76 \\ 
    & One-Shot & \gray{68}0.85 & \gray{72}0.87 & \gray{62}0.52 & \gray{65}0.52 & \gray{67}0.07 & \gray{55}0.78 \\ 
    & Two-Shot & \gray{50}0.79 & \gray{52}0.79 & \gray{46}0.42 & \gray{46}0.40 & \gray{70}0.08 & \gray{57}0.79 \\ 
    & Three-Shot & \gray{48}0.78 & \gray{50}0.78 & \gray{62}0.52 & \gray{65}0.52 & \gray{76}0.08 & \gray{60}0.79 \\ 
    & Fine-Tuned & \gray{98}\textbf{0.97 }& \gray{99}0.97 & \gray{96}0.73 & \gray{95}0.71 & \gray{64}0.07 & \gray{100}\textbf{0.85} \\ 
    \hline 
    \hline
    \multirow{5}{*}{Mistral-7B-Instruct} & Zero-Shot & \gray{61}0.83 & \gray{70}0.86 & \gray{79}0.62 & \gray{81}0.62 & \gray{51}0.06 & \gray{44}0.77 \\ 
    & One-Shot & \gray{23}0.69 & \gray{16}0.64 & \gray{79}0.62 & \gray{81}0.62 & \gray{66}0.07 & \gray{56}0.79 \\ 
    & Two-Shot & \gray{11}0.64 & \gray{0}0.58 & \gray{59}0.50 & \gray{67}0.53 & \gray{62}0.07 & \gray{62}0.79 \\ 
    & Three-Shot & \gray{60}0.82 & \gray{64}0.84 & \gray{74}0.59 & \gray{79}0.61 & \gray{65}0.07 & \gray{64}0.80 \\ 
    & Fine-Tuned & \gray{100}\textbf{0.97} & \gray{100}\textbf{0.98} & \gray{100}\textbf{0.75} & \gray{100}\textbf{0.74} & \gray{100}\textbf{0.10} & \gray{70}0.81 \\ 
    \hline
    \multirow{5}{*}{Llama-2-7B-Chat} & Zero-Shot & \gray{22}0.68 & \gray{52}0.79 & \gray{19}0.25 & \gray{1}0.11 & \gray{0}0.02 & \gray{0}0.70 \\ 
    & One-Shot & \gray{1}0.60 & \gray{42}0.75 & \gray{19}0.25 & \gray{1}0.11 & \gray{46}0.06 & \gray{39}0.76 \\ 
    & Two-Shot & \gray{2}0.60 & \gray{43}0.75 & \gray{17}0.24 & \gray{0}0.10 & \gray{52}0.06 & \gray{55}0.78 \\ 
    & Three-Shot & \gray{46}0.77 & \gray{54}0.80 & \gray{17}0.24 & \gray{0}0.10 & \gray{46}0.06 & \gray{57}0.79 \\ 
    & Fine-Tuned & \gray{100}\textbf{0.97} & \gray{100}\textbf{0.98} & \gray{98}0.74 & \gray{98}0.73 & \gray{70}0.08 & \gray{67}0.80 \\ 
    \hline
    \multirow{5}{*}{Falcon-7B-Instruct} & Zero-Shot & \gray{0}0.59 & \gray{43}0.75 & \gray{53}0.46 & \gray{56}0.46 & \gray{56}0.07 & \gray{55}0.78 \\ 
    & One-Shot & \gray{0}0.59 & \gray{37}0.73 & \gray{16}0.23 & \gray{29}0.29 & \gray{20}0.04 & \gray{15}0.73 \\ 
    & Two-Shot & \gray{1}0.60 & \gray{43}0.75 & \gray{4}0.16 & \gray{3}0.13 & \gray{34}0.05 & \gray{26}0.74 \\ 
    & Three-Shot & \gray{1}0.60 & \gray{43}0.75 & \gray{4}0.16 & \gray{3}0.12 & \gray{19}0.04 & \gray{22}0.74 \\ 
    & Fine-Tuned & \gray{96}0.96 & \gray{97}0.97 & \gray{96}0.73 & \gray{96}0.72 & \gray{87}0.09 & \gray{55}0.78 \\ 
    \hline 
    \end{tabular}
    \caption{\footnotesize Performance for three open and two closed models on two classification tasks and one text summarization task. GPT-4 outperforms other models in few-shot settings, but open models are competitive after fine-tuning with modest assumptions. \looseness=-1}
    \label{tab:main_results}
\end{table*}

\section{Multifaceted Evaluation of Open vs. Closed Models}

We select a practical, representative sample of tasks, including those that 1) reflect real-world uses of generative instruction-tuned models (\textit{e.g.}, fact-checking chatbots, like Aos Fatos’ FatimaGPT \cite{aosfatos} or Meedan’s Check \cite{meedan}); and 2) reflected consequential work envisioned by other research. For example, \citet{gilardi2023chatgpt} suggest ChatGPT can be used for data annotation (we consider specifically entity resolution), and \citet{waisberg2023gpt} explore GPT-4 for triaging patients via clinical dialogues. We acknowledge it may not be \textit{desirable} to use an LM in a setting like clinical dialogue summarization or fact-checking, especially without human supervision, and that our tasks are \textit{proxies} to real-world applications. \looseness=-1

\subsection{Representative General Tasks}

We study three tasks to compare performance of open vs. closed models, with sample and evaluation metrics in Table \ref{tab:performance_tasks}. \looseness=-1

\begin{enumerate}[leftmargin=6mm]
    \item \textbf{Entity Resolution}: We use a custom dataset of public records to evaluate performance on Entity Resolution \cite{10.14778/2367502.2367564}. Given two pairs of names and addresses, the model determines whether the pairs refer to the same person. One set is derived from home deeds in Mecklenburg County, NC; the other comes from voter records. The dataset contains 1,000 records annotated by three humans (Krippendorff's $\alpha$ of 0.88, 95\% CI: 0.85, 0.90 \cite{zapf_measuring_2016}). \looseness=-1 
    \item \textbf{Fact-Checking}: We use the Climate-FEVER dataset \cite{diggelmann2020climate} to evaluate performance on a fact-checking task. Given a climate-related claim and an associated piece of evidence, the model answers whether the evidence Supports, Refutes, or provides insufficient information to support or refute the claim \cite{diggelmann2020climate}. For predefined training, validation, and test splits, we use the version of this dataset available at \url{https://huggingface.co/datasets/amandakonet/climate_fever_adopted}, used in fine-tuning in-domain climate fact-checking models like a \href{https://huggingface.co/amandakonet/climatebert-fact-checking}{Climate-BERT} \cite{webersinke2021climatebert}. 
    \item \textbf{Clinical Dialogue Summarization}: We use the MTS-Dialog dataset \cite{mts-dialog} to evaluate models on clinical dialogue summarization, following prior work \cite{yim2023overview,han2023huskyscribe}. Given a dialogue between doctor and patient, plus the topic (\textit{e.g.,} medication history, chief complaint), the model must summarize the dialogue, capturing information relevant to the topic. \looseness=-1
\end{enumerate}

\noindent A simple postprocessing script removed extra words so model output could be measured against labels for tasks 1-2. Given ``The answer is Supports'' for fact-checking, the script removes ``The answer is.'' \looseness=-1

\subsection{Performance --- Fine-tuned Open Models Can Outperform Closed Models}

As shown in Table \ref{tab:main_results}, GPT-4-Turbo outperforms open models in the few-shot setting, and by substantial margins for the entity resolution and fact-checking tasks. Of the open models, only Mistral-7B-Instruct is competitive with GPT-3.5-Turbo in the few-shot setting. Fine-tuning for a single dataset epoch, however, yields open models that are competitive and in some cases even outperform GPT-4-Turbo and fine-tuned GPT-3.5-Turbo. LLaMA-2-7B-Chat achieves no more than 25\% accuracy on the fact-checking task in any few-shot setting, yet outperforms GPT-4-Turbo after fine-tuning. GPT-4-Turbo also achieves the best few-shot performance on medical summarization. With fine-tuning, though, Mistral-7B-Instruct outperforms GPT-4-Turbo few shot, achieving higher BLEU score (but not higher BERT score) than GPT-3.5-Turbo, while fine-tuned LLaMA-2-7B-Chat and Falcon-7B-Instruct achieve results competitive with few-shot GPT-4-Turbo. \looseness=-1

\subsection{Cost Analysis --- Open Models Are More Affordable}

To better understand the financial cost of customizing and using open models versus using closed models out of the box, we compute the approximate cost of inference and of fine-tuning for the climate fact-checking task. For closed models, we obtain the number of input tokens in our test dataset using the \href{https://github.com/openai/tiktoken}{tiktoken tokenizer} for OpenAI models. We multiply this total by the per-token costs \href{https://openai.com/pricing}{published by OpenAI}. We omit the cost of output tokens in this computation, which we estimate to be less than 1\% of the total cost of inference for our tasks. We compute cost for open models by taking the per-hour price of our cloud instance times the runtime logged to our Weights and Biases \cite{wandb} account. Costs reported are consistent with billing by OpenAI and GCP. We also report runtime for open and closed models. \looseness=-1

\begin{table*}[htbp]
    \centering
    \footnotesize
    \begin{tabular}{llccccc}
    \toprule
    Model & Scenario & Input Tokens & 1k Token Cost & Runtime Hours & Cloud Cost & Total Cost \\
    \midrule
\multirow{4}{*}{GPT4-Turbo} & Zero-Shot & 260,056 & \$0.010 & \gray{1}0.32 & N/A & \gray{36}\$2.60 \\
    & One-Shot & 385,926 & \$0.010 & \gray{2}0.34 & N/A & \gray{56}\$3.86 \\
    & Two-Shot & 484,166 & \$0.010 & \gray{1}0.31 & N/A & \gray{72}\$4.84 \\
    & Three-Shot & 550,171 & \$0.010 & \gray{1}0.32 & N/A & \gray{82}\$5.50 \\
    \hline
    \multirow{5}{*}{GPT3.5-Turbo} & Zero-Shot & 260,056 & \$0.001 & \gray{0}0.20 & N/A & \gray{0}\$0.26 \\
    & One-Shot & 385,926 & \$0.001 & \gray{1}0.23 & N/A & \gray{1}\$0.39 \\
    & Two-Shot & 484,166 & \$0.001 & \gray{0}0.20 & N/A & \gray{3}\$0.48 \\
    & Three-Shot & 550,171 & \$0.001 & \gray{0}0.20 & N/A & \gray{4}\$0.55 \\
    & Fine-Tuning & 260,056 & \$0.003 & \gray{12}1.54 & N/A & \gray{100}\$6.60 \\
    & Fine-Tuned & 260,056 & \$0.003 & \gray{0}0.11 & N/A & \gray{8}\$0.78 \\
    \hline
    \multirow{5}{*}{Falcon-7B-Instruct} & Zero-Shot & N/A & N/A & \gray{6}0.84 & \$0.32 & \gray{0}\$0.27 \\
    & One-Shot & N/A & N/A & \gray{10}1.24 & \$0.32 & \gray{2}\$0.40 \\
    & Two-Shot & N/A & N/A & \gray{11}1.37 & \$0.32 & \gray{2}\$0.44 \\
    & Three-Shot & N/A & N/A & \gray{13}1.58 & \$0.32 & \gray{3}\$0.50 \\
    & Fine-Tuning & N/A & N/A & \gray{91}9.95 & \$0.32 & \gray{46}\$3.18 \\
    & Fine-Tuned & N/A & N/A & \gray{7}0.96 & \$0.32 & \gray{0}\$0.31 \\
    \hline
    \multirow{5}{*}{LLaMA-2-7B-Chat} & Zero-Shot & N/A & N/A & \gray{7}0.91 & \$0.32 & \gray{0}\$0.29 \\
    & One-Shot & N/A & N/A & \gray{10}1.27 & \$0.32 & \gray{2}\$0.41 \\
    & Two-Shot & N/A & N/A & \gray{12}1.48 & \$0.32 & \gray{3}\$0.47 \\
    & Three-Shot & N/A & N/A & \gray{14}1.64 & \$0.32 & \gray{4}\$0.53 \\
    & Fine-Tuning & N/A & N/A & \gray{100}10.92 & \$0.32 & \gray{50}\$3.49 \\
    & Fine-Tuned & N/A & N/A & \gray{8}1.08 & \$0.32 & \gray{1}\$0.34 \\
    \hline
    \multirow{5}{*}{Mistral-7B-Instruct} & Zero-Shot & N/A & N/A & \gray{8}0.97 & \$0.32 & \gray{0}\$0.31 \\
    & One-Shot & N/A & N/A & \gray{9}1.10 & \$0.32 & \gray{1}\$0.35 \\
    & Two-Shot & N/A & N/A & \gray{10}1.25 & \$0.32 & \gray{2}\$0.40 \\
    & Three-Shot & N/A & N/A & \gray{11}1.37 & \$0.32 & \gray{2}\$0.44 \\
    & Fine-Tuning & N/A & N/A & \gray{99}10.90 & \$0.32 & \gray{50}\$3.49 \\
    & Fine-Tuned & N/A & N/A & \gray{7}0.92 & \$0.32 & \gray{0}\$0.29 \\
    \bottomrule
    \end{tabular}    
    \caption{\footnotesize Open models are less costly than GPT-4-Turbo, based on costs computed using fact-checking data. The cost of fine-tuning GPT-3.5-Turbo includes 727,845 Training Tokens, billed at \$0.008 per 1,000. \looseness=-1}
    \label{tab:cost_runtime}
\end{table*}

If laboratory-scale AI is feasible, we expect open models to be cost-competitive with closed models, and ideally more affordable. Table \ref{tab:cost_runtime} shows that the few-shot cost of GPT-4-Turbo is approximately ten times that of a few-shot open model or GPT-3.5.-Turbo. The cost of fine-tuning any open model for one dataset epoch and evaluating it once (``Fine-Tuning'' in Table \ref{tab:cost_runtime}), a process which Performance results indicate produces a superior fact-checking model to GPT-4-Turbo, is lower than the cost of running inference once using GPT-4-Turbo in the one-shot setting. The most significant savings come when using the model after fine-tuning (``Fine-Tuned'' in Table \ref{tab:cost_runtime}). Fine-tuned open models are much less expensive than GPT-4-Turbo, and more performant than few-shot closed models. \looseness=-1

Closed models excel on runtime. Fine-tuned GPT-3.5-Turbo is the fastest option, and ten times faster than open models. Few-shot GPT-4-Turbo requires 1.5 times as long as few-shot GPT-3.5-Turbo, but is three times as fast as open models. Our measurements do not include all costs, such as purchasing persistent disk storage, static IPs, and more reliable cloud instances, but provides an empirically grounded analysis of the cost of entry to locally train and deploy a model. \looseness=-1

\subsection{Data Responsiveness --- Modest Fine-tuning Can Make Open Models Competitive}

To understand the amount of data needed to produce a domain-specific open model, we study the performance of LLaMA-2-7B-Chat checkpoints for clinical dialogue summarization, entity resolution, and climate fact-checking tasks. We save intermediate model weights at 20\%, 40\%, 60\%, 80\%, and 100\% of each task-specific training dataset, and assess the intermediate model on the full test dataset. Moreover, for the climate fact-checking task, which has a larger training set of 4,298 samples, we save checkpoints every 500 samples, and assess accuracy using these checkpoints on 150 test samples (approximately 10\% of the test dataset). We save these 500-step fact-checking checkpoints for LLaMA-2-7B-Chat; Mistral-7B-Instruct; Falcon-7B-Instruct; and GPT-3.5-Turbo. Because OpenAI does not allow saving model checkpoints during fine-tuning, we submit separate fine-tuning jobs for GPT-3.5-Turbo using subsets of the training dataset. \looseness=-1

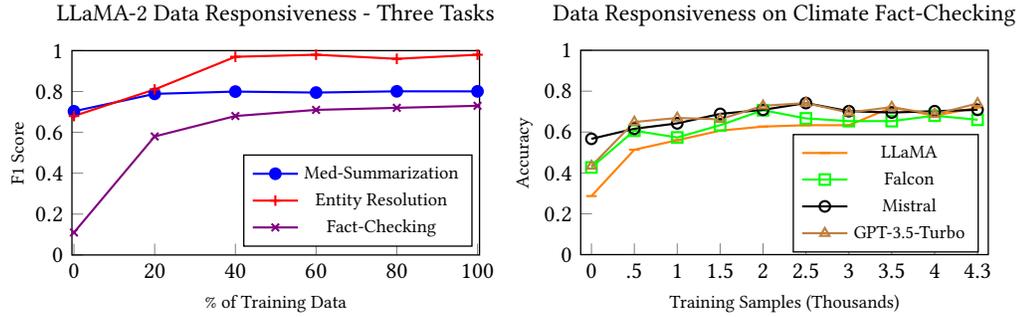
\begin{figure}[!ht]
\begin{tikzpicture}
\begin{axis} [
    height=4.3cm,
    width=7cm,
    line width = .5pt,
    ymin = 0, 
    ymax = 1,
    xmin=-.5,
    xmax=100.5,
    ylabel={F1 Score},
    ylabel shift=-2pt,
    xtick = {0,20,40,60,80,100},
    xticklabels = {0,20,40,60,80,100},
    xtick pos=left,
    ytick pos = left,
    xlabel style = {font=\footnotesize},
    ylabel style = {font=\footnotesize},
    title=LLaMA-2 Data Responsiveness - Three Tasks,
    xlabel= {\% of Training Data},
    legend style={at={(.42,.04)},anchor=south west,font=\footnotesize}
]

\addplot[thick,solid,mark=*,color=blue] coordinates {(0,0.7032) (20,0.7885) (40,0.7996) (60,0.7948) (80,0.8009) (100,0.8007)};
\addplot[thick,solid,mark=+,color=red] coordinates {(0,0.68) (20,0.81) (40,0.97) (60,0.98) (80,0.96) (100,0.98)};
\addplot[thick,solid,mark=x,color=violet] coordinates {(0,0.11) (20,0.58) (40,0.68) (60,0.71) (80,0.72) (100,0.73)};

\legend {Med-Summarization, Entity Resolution, Fact-Checking};
\end{axis}
\end{tikzpicture}
\begin{tikzpicture}
\begin{axis} [
    height=4.3cm,
    width=7cm,
    line width = .5pt,
    ymin = 0, 
    ymax = 1,
    xmin=-.25,
    xmax=9.25,
    ylabel={Accuracy},
    ylabel shift=-2pt,
    xtick = {0,1,2,3,4,5,6,7,8,9},
    xticklabels = {0,.5, 1, 1.5, 2, 2.5, 3, 3.5, 4, 4.3},
    xtick pos=left,
    ytick pos = left,
    title=Data Responsiveness on Climate Fact-Checking,
    xlabel= {Training Samples (Thousands)},
    xlabel style = {font=\footnotesize},
    ylabel style = {font=\footnotesize},
    legend style={at={(.52,.01)},anchor=south west, font=\footnotesize}
]

\addplot[thick,solid,mark=-,color=orange] coordinates {(0, 0.2866666666666667) (1, 0.5133333333333333) (2, 0.56) (3, 0.6066666666666667) (4, 0.6266666666666667) (5, 0.6333333333333333) (6, 0.6333333333333333) (7, 0.72) (8, 0.68) (9, 0.72)};
\addplot[thick,solid,mark=square,color=green] coordinates {(0, 0.4266666666666667) (1, 0.6066666666666667) (2, 0.5733333333333334) (3, 0.6333333333333333) (4, 0.7066666666666667) (5, 0.6666666666666666) (6, 0.6533333333333333) (7, 0.6533333333333333) (8, 0.68) (9, 0.66)};
\addplot[thick,solid,mark=o,color=black] coordinates {(0, 0.5666666666666667) (1, 0.6158940397350994) (2, 0.6423841059602649) (3, 0.6887417218543046) (4, 0.7086092715231788) (5, 0.7417218543046358) (6, 0.7019867549668874) (7, 0.695364238410596) (8, 0.7019867549668874) (9, 0.7086092715231788)};
\addplot[thick,solid,mark=triangle,color=brown] coordinates {(0,0.43333333333333335)(1, 0.6490066225165563) (2, 0.6688741721854304) (3, 0.6622516556291391) (4, 0.7284768211920529) (5, 0.7417218543046358) (6, 0.695364238410596) (7, 0.7218543046357616) (8, .6866666666666666) (9, .74)};

\legend {LLaMA, Falcon, Mistral, GPT-3.5-Turbo};
\end{axis}
\end{tikzpicture}
\caption{\footnotesize Left: Fine-tuning improvements emerge during the first 50\% of the training data, only a few hundred training samples in the case of Medical Summarization and Entity Resolution. Right: Finetuned open models are competitive with finetuned GPT-3.5-Turbo with little data (1,000 fact-checking samples). \looseness=-1}
\label{fig:bothresponsiveness}
\end{figure}

\noindent If laboratory-scale AI is feasible, we would expect that massive data-gathering projects would not be needed to produce a competitive in-domain model. As shown in Figure \ref{fig:bothresponsiveness} (left plot), LLaMA-2-Chat-7B achieves BERTScore-F1 of .79 on clinical dialogue summarization after only 20\% of training samples (240 samples), and .97 F1 on entity resolution after only 40\% of training samples. Similarly (right plot), Mistral-7B-Instruct trained on climate fact-checking achieves accuracy of .71 after 2,000 samples, while LLaMA-2-Chat-7B achieves accuracy comparable to fine-tuned GPT-3.5-Turbo after about 3,500 samples. Fine-tuned laboratory-scale models capable of results comparable to GPT-4-Turbo can be trained using quantities of data feasible for researchers to gather. Variance among open models reflects base model benchmark performance, with Mistral generally outperforming LLaMA-2, and LLaMA-2 outperforming Falcon \cite{liang2023holistic,hendrycks2021measuring}, suggesting pretraining disparities (\textit{e.g.}, LLaMA-2 pretraining on a larger dataset than Falcon) carry over during domain adaptation. \looseness=-1

\subsection{Model Generality --- Fine-tuning Does Not Inhibit the Generality of Open Models}

While fine-tuning may improve the performance of a chat-based model for a specific task, it is not clear whether this would compromise the model's general-purpose utility when a user interacts with it via natural language. To study whether the model maintains this utility, we evaluate each of the domain-specific (entity resolution, fact-checking, and clinical dialogue summarization) LLaMA-2-Chat-7B models on the other tasks for which the model was not fine-tuned. We then compare the domain-specific model's performance on each task against the general-purpose base LLaMA-2. \looseness=-1 

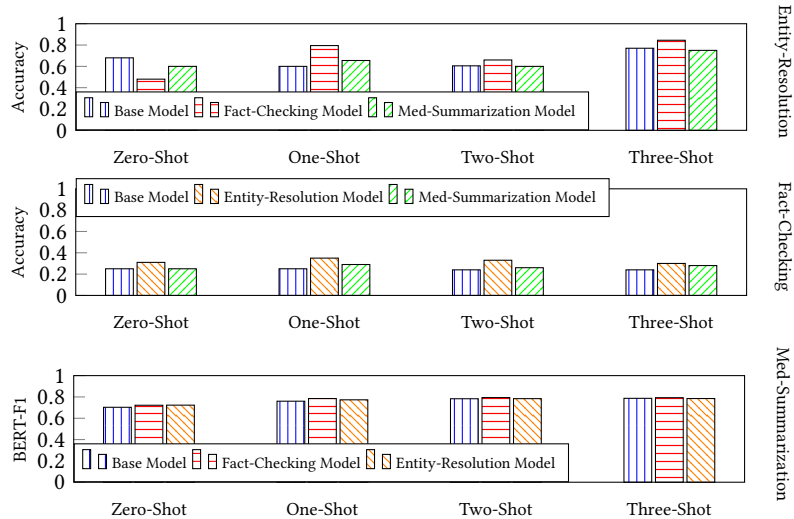
\begin{figure*}[!htbp]
\begin{tikzpicture}
    \begin{axis} [
        height=3cm,
        width=10.5cm,
        ybar = .05cm,
        bar width = 10.5pt,
        ymin = 0, 
        ymax = 1,
        ylabel=Accuracy,
        ylabel style={font=\footnotesize},
        ylabel shift=-2pt,
        xtick = {1,2,3,4},
        xtick style={draw=none},
        ytick pos = left,
        xticklabels = {Zero-Shot, One-Shot, Two-Shot, Three-Shot},
        title=Entity-Resolution,
        xticklabel style={font=\footnotesize},
        x label style={at={(axis description cs:0.5,-0.25)},anchor=north,font=\footnotesize},
        enlarge x limits={abs=1cm},
        legend style={at={(0,0)},anchor=south west,font=\scriptsize, legend columns=-1},
        title style={at={(1.08,0.4)},anchor=north, rotate=-90, font=\footnotesize},
    ]    
    \addplot [pattern=vertical lines,pattern color = blue] coordinates {(1,.68) (2,.6) (3,.605) (4,.77)};    
    \addplot [pattern=horizontal lines,pattern color = red] coordinates {(1,.48) (2,.795) (3,.66) (4,.845)};    
    \addplot [pattern=north east lines,pattern color = green] coordinates {(1,.60) (2,.655) (3,.60) (4,.750)};   
    \legend {Base Model, Fact-Checking Model, Med-Summarization Model};    
    \end{axis}
\end{tikzpicture}

\begin{tikzpicture}
    \begin{axis} [
        height=3cm,
        width=10.5cm,
        ybar = .05cm,
        bar width = 10.5pt,
        ymin = 0, 
        ymax = 1,
        ylabel=Accuracy,
        ylabel style={font=\footnotesize},
        ylabel shift=-2pt,
        xtick = {1,2,3,4},
        xtick style={draw=none},
        ytick pos = left,
        xticklabels = {Zero-Shot, One-Shot, Two-Shot, Three-Shot},
        title= Fact-Checking,
        xticklabel style={font=\footnotesize},
        x label style={at={(axis description cs:0.5,-0.25)},anchor=north,font=\footnotesize},
        enlarge x limits={abs=1cm},
        legend style={at={(0,0.735)},anchor=south west,font=\scriptsize,legend columns=-1},
        title style={at={(1.08,0.4)},anchor=north, rotate=-90, font=\footnotesize},
    ]    
    \addplot [pattern=vertical lines,pattern color = blue] coordinates {(1,.25) (2,.25) (3,.24) (4,.24)};    
    \addplot [pattern=north west lines,pattern color = orange] coordinates {(1,.31) (2,.35) (3,.33) (4,.30)};    
    \addplot [pattern=north east lines,pattern color = green] coordinates {(1,.25) (2,.29) (3,.26) (4,.28)};    
    \legend {Base Model, Entity-Resolution Model, Med-Summarization Model};    
    \end{axis}
\end{tikzpicture}

\begin{tikzpicture}
    \begin{axis} [
        height=3cm,
        width=10.5cm,
        ybar = .05cm,
        bar width = 10.5pt,
        ymin = 0, 
        ymax = 1,
        ylabel=BERT-F1,
        ylabel style={font=\footnotesize},
        ylabel shift=-2pt,
        xtick = {1,2,3,4},
        xtick style={draw=none},
        ytick pos = left,
        xticklabels = {Zero-Shot, One-Shot, Two-Shot, Three-Shot},
        title= Med-Summarization,
        xticklabel style={font=\footnotesize},
        x label style={at={(axis description cs:0.5,-0.25)},anchor=north,font=\footnotesize},
        enlarge x limits={abs=1cm},
        legend style={at={(0,0)},anchor=south west,font=\scriptsize,legend columns=-1},
        title style={at={(1.08,0.4)},anchor=north, rotate=-90, font=\footnotesize},
    ]    
    \addplot [pattern=vertical lines,pattern color = blue] coordinates {(1,.703) (2,.760) (3,.783) (4,.787)};    
    \addplot [pattern=horizontal lines,pattern color = red] coordinates {(1,.722) (2,.785) (3,.794) (4,.793)};   
    \addplot [pattern=north west lines,pattern color = orange] coordinates {(1,.723) (2,.773) (3,.784) (4,.785)}; 
    \legend {Base Model, Fact-Checking Model, Entity-Resolution Model};
    \end{axis}
\end{tikzpicture}

\caption{\footnotesize Models fine-tuned on a task using qLoRA offer strong zero-shot performance on other tasks, often stronger than the base model.}
\label{fig:model_generality_two}
\end{figure*}
If the model maintains a general purpose utility, we would expect to see at worst insignificant decreases in the performance of a fine-tuned model when compared with the base model. As illustrated in Figure \ref{fig:model_generality_two}, performance actually increases marginally in most cases when using fine-tuned models on tasks for which they were not fine-tuned. For example, the fine-tuned fact-checking model exceeds base model performance in the one, two, and three shot settings for the entity resolution task. This may not mean that low-rank fine-tuning will always improve performance on related tasks, but our findings suggest that fine-tuning for a specific domain does not degrade the general-purpose utility of an open model. \looseness=-1

\section{Responsible Use of Open Models}

One of the presumed advantages offered by closed models is the process used to mitigate bias and prevent the closed model from generating harmful or inaccurate output. We thus evaluate three scenarios related to responsible and transparent model use: question answering under differential privacy (privacy), toxicity classification (bias), and abstention, referring to a model refusing to confidently answer questions for which it does not have the answer (transparency). \looseness=-1

\subsection{Differential Privacy --- Privately Fine-tuned Open Models Approach Non-Private Performance}
Differentially private (DP) deep learning (using a privatized gradient descent optimizer \cite{abadi2016deep}) has been adopted to protect users and avoid legal risks of sensitive data use \cite{gong2020survey, ghazi2021deep}. While challenging in the context of language models \cite{peris2023privacy}, recent work \cite{yu2021differentially, dp-transformers} demonstrates the potential to train general purpose models using differentially private fine-tuning on sensitive data \cite{yousefpour2021opacus}. We adopt the perspective of a small medical lab with sensitive data, seeking to privately fine-tune an open-source, general purpose medical model. We use the MedQA \cite{jin2021disease} task as a proxy for this scenario (included in the MultiMedBench \cite{tu2023towards} benchmark), simplifying it to a binary classification task. We employ private fine-tuning with qLoRA, and report results at five levels of privacy ($\epsilon$ = 0.5, 1, 5, 20, $\infty$, where lower $\epsilon$ denotes greater privacy, and $\epsilon=\infty$ is non-private).  \looseness=-1

\begin{table*}[htbp]
    \centering
    \small
    \tabcolsep=0.14cm
    \renewcommand*{\arraystretch}{1.1}
    \begin{tabular}{|l|l||c|c|c|c|c||c|c|c|c|c|}
    \hline
    \multirow{2}{*}{Scenario} & \multirow{2}{*}{Model (finetuned)} & \multicolumn{5}{|c||}{Acc. at Privacy Level} & \multicolumn{5}{|c|}{F1 at Privacy Level} \\
    \cline{3-12}
    & & $\epsilon = 0.5$ & $\epsilon = 1$ & $\epsilon = 5$ & $\epsilon = 20$ & $\epsilon = \infty$ & $\epsilon = 0.5$ & $\epsilon = 1$ & $\epsilon = 5$ & $\epsilon = 20$ & $\epsilon = \infty$ \\
     \hline 
    \multirow{3}{*}{MedQA-TF} & Falcon-7B-Instruct & \gray{74}0.47 & \gray{59}0.51 & \gray{0}0.52 & \gray{14}0.53 & \gray{0}0.52 & \gray{33}0.35 & \gray{0}0.36 & \gray{0}0.36 & \gray{0}0.36 & \gray{0}0.51 \\ 
    & Llama-2-7B-Chat & \gray{0}0.19 & \gray{0}0.41 & \gray{0}0.52 & \gray{0}0.52 & \gray{31}0.56 & \gray{0}0.26 & \gray{53}0.46 & \gray{85}0.53 & \gray{78}0.54 & \gray{29}0.55 \\ 
    & Mistral-7B-Instruct & \gray{100}0.57 & \gray{100}0.58 & \gray{100}0.59 & \gray{100}0.59 & \gray{100}0.65 & \gray{100}0.53 & \gray{100}0.55 & \gray{100}0.56 & \gray{100}0.59 & \gray{100}0.65 \\ 
    \hline 
    \end{tabular}
    \caption{\footnotesize Privately tuned models can approach non-private performance at lower levels of privacy.}
    \label{tab:private_results}
\end{table*}

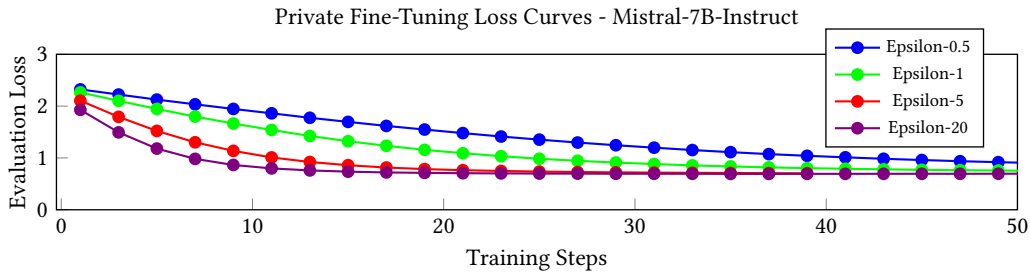
\begin{figure}[htbp]
\begin{tikzpicture}
\begin{axis} [
    height=3.65cm,
    width=.95\linewidth,
    line width = .5pt,
    ymin = 0, 
    ymax = 3, 
    xmin=-.25,
    xmax=50,
    ylabel={Evaluation Loss},
    ylabel shift=-2pt,
    xtick = {0,10,20,30,40,50},
    xticklabels = {0,10,20,30,40,50},
    xtick pos=left,
    ytick pos = left,
    title=Private Fine-Tuning Loss Curves - Mistral-7B-Instruct,
    xlabel= {Training Steps},
    legend style={at={(.8,.4)},anchor=south west,font=\footnotesize},
]
\addplot[thick,solid,mark=*,color=blue] coordinates {
    (1, 2.32259082794189) (3, 2.22330260276794) (5, 2.12590336799622) (7, 2.03487610816956) (9, 1.94560348987579) (11, 1.86100566387177) (13, 1.77480590343475) (15, 1.69581258296967) (17, 1.61793041229248) (19, 1.54781365394592) (21, 1.47933447360992) (23, 1.41326224803925) (25, 1.35318303108215) (27, 1.29711079597473) (29, 1.24528205394745) (31, 1.19834756851196) (33, 1.15241539478302) (35, 1.11162877082825) (37, 1.07390427589417) (39, 1.04121947288513) (41, 1.01199626922607) (43, 0.983536005020142) (45, 0.959699511528015) (47, 0.937240183353424) (49, 0.918206214904785) (51, 0.900966465473175) (53, 0.884782075881958) (55, 0.870853662490845) (57, 0.857222497463226)
};

\addplot[thick,solid,mark=*,color=green] coordinates {
    (1, 2.26467752456665) (3, 2.10215497016907) (5, 1.94571697711945) (7, 1.7984961271286) (9, 1.66442930698395) (11, 1.5401371717453) (13, 1.42434811592102) (15, 1.32256054878235) (17, 1.23409235477448) (19, 1.15640640258789) (21, 1.09133422374725) (23, 1.03361618518829) (25, 0.984456241130829) (27, 0.944549679756165) (29, 0.911096453666687) (31, 0.882067501544952) (33, 0.856262505054474) (35, 0.836006581783295) (37, 0.817877948284149) (39, 0.80299723148346) (41, 0.790962636470795) (43, 0.780539751052856) (45, 0.770221650600433) (47, 0.762710750102997) (49, 0.757533073425293) (51, 0.750719010829926) (53, 0.747323572635651) (55, 0.741247355937958) (57, 0.737527310848236)
};

\addplot[thick,solid,mark=*,color=red] coordinates {
    (1, 2.10511350631714) (3, 1.79268217086792) (5, 1.52101755142212) (7, 1.30202984809875) (9, 1.1365419626236) (11, 1.00991332530975) (13, 0.921977281570435) (15, 0.85982072353363) (17, 0.814881086349487) (19, 0.784619033336639) (21, 0.76365739107132) (23, 0.748596251010895) (25, 0.737012028694153) (27, 0.72871196269989) (29, 0.72230452299118) (31, 0.716553032398224) (33, 0.713071584701538) (35, 0.709501028060913) (37, 0.706278860569) (39, 0.703549027442932) (41, nan) (43, nan) (45, nan) (47, nan) (49, nan) (51, nan) (53, nan) (55, nan) (57, nan)
};

\addplot[thick,solid,mark=*,color=violet] coordinates {
    (1, 1.92817461490631) (3, 1.49368846416473) (5, 1.18067789077759) (7, 0.982162892818451) (9, 0.86459892988205) (11, 0.798163592815399) (13, 0.759200692176819) (15, 0.735446512699127) (17, 0.720092117786408) (19, 0.711881816387177) (21, 0.705342769622803) (23, 0.701900243759155) (25, 0.69851690530777) (27, 0.697536170482636) (29, 0.695481717586517) (31, 0.695319354534149) (33, 0.694554567337036) (35, 0.692434132099152) (37, 0.692108988761902) (39, 0.69288432598114) (41, 0.693157553672791) (43, 0.69385039806366) (45, 0.694264829158783) (47, 0.693800330162048) (49, 0.695092439651489) (51, 0.696992993354797) (53, 0.697064876556397) (55, 0.698918163776398) (57, 0.700154066085815)
};

\legend {Epsilon-0.5, Epsilon-1, Epsilon-5, Epsilon-20};
\end{axis}
\end{tikzpicture}
\caption{\footnotesize Increasing privacy (by decreasing $\epsilon$) leads to noisier gradients, delaying convergence; but privately trained open models do learn.}
\label{fig:private_loss}
\end{figure}

\noindent Table~\ref{tab:private_results} illustrates how challenging MedQA-TF proved for open models, which performed much lower than the state-of-the-art \cite{tu2023towards}. However, our results show that private fine-tuning allowed a model like Mistral-7B-Instruct to approach its non-privately fine-tuned performance at $\epsilon$=20. Figure~\ref{fig:private_loss} demonstrates how different privacy settings used in Mistral-7B-Instruct impact evaluation loss curves, showing that at lower $\epsilon$, models take longer to converge. A challenge of noisy, privatized updates is that batch size needs to be large, posing issues for lab-scale approaches that use smaller batches.  \looseness=-1

\subsection{Toxicity Bias --- Open Models Improve with Fine-Tuning, But Lag Behind Closed Models}

We evaluate open and closed models on a subset of CivilComments-WILDS~\cite{koh2021wilds}, a dataset of real online comments curated from the Civil Comments platform. Dataset labels describe the toxicity of the comment and whether a demographic membership is mentioned in the comment. Models must classify whether a comment is toxic, and their classifications are analyzed through the lens of performance and fairness (whether classifications are incorrect more often for certain demographic groups). We report 1) accuracy on all comments assessed and 2) worst-group accuracy, which represents the lowest accuracy after segmenting the model's output by demographic membership and toxicity label (\textit{e.g.,} worst-group accuracy might refer to accuracy for non-toxic comments and male demographic membership). To ensure a controlled and interpretable experiment, we limited the demographic groups to Male and Female, such that the measurements correspond to gender bias. We used 800 training, 100 validation, and 200 test samples from the dataset. Training, validation, and test data were balanced across the four groups (Male Toxic, Male Non-Toxic, Female Toxic, Female Non-Toxic). \looseness=-1

As shown in Table~\ref{tab:bias_f1}, closed models outperform open models on this assessment. Fine-tuning improves overall (Mean) accuracy for Mistral and Falcon, but had no discernable effect for LLaMA-2. Fine-tuning did not increase Worst-Group accuracy over performance in the few-shot setting for any of the open models. The strongest performing model of the group was three-shot GPT-4-Turbo, which exceeded other models in both Mean and Worst-Group accuracy. Fine-tuned GPT-3.5-Turbo matches three-shot GPT-4-Turbo on overall accuracy, but not Worst-Group accuracy. However, the task is difficult, and three-shot Mistral-7B-Instruct surprisingly outperforms zero-shot GPT-4-Turbo on Worst-Group accuracy. \looseness=-1

\begin{table*}[htbp]
    \centering
    \footnotesize
    \begin{tabular}{|c | c c | c c | c c | c c | c c|}
    \hline
    \multirow{2}{*}{Scenario} & \multicolumn{2}{c|}{GPT-4-Turbo} & \multicolumn{2}{c|}{GPT-3.5-Turbo} & \multicolumn{2}{c|}{Mistral-7B-Instruct} & \multicolumn{2}{c|}{Falcon-7B-Instruct} & \multicolumn{2}{c|}{Llama-2-7B-Chat}\\
    \cline{2-11}
    & Worst(\%) & Mean(\%) & Worst(\%) & Mean(\%) & Worst(\%) & Mean(\%) & Worst(\%) & Mean(\%) & Worst(\%) & Mean(\%) \\ 
    \hline
    Zero-Shot&\gray{40}0.37 & \gray{56}0.63 & \gray{68}0.61 & \gray{68}0.66 & \gray{0}0.1 & \gray{50}0.53 & \gray{0}0 & \gray{50}0.51 & \gray{0}0.14 & \gray{50}0.52 \\
    One-Shot&\gray{40}0.37 &\gray{70}0.64 & \gray{55}0.59 & \gray{58}0.63 & \gray{30}0.3 & \gray{50}0.49 & \gray{40}0.4 & \gray{50}0.5 & \gray{0}0.1 & \gray{50}0.5\\
    Two-Shot&\gray{77}0.63 &\gray{77}0.68 & \gray{60}0.62 & \gray{60}0.64 & \gray{40}0.41 & \gray{50}0.51 & \gray{50}0.49 & \gray{50}0.53 & \gray{10}0.14 & \gray{50}0.52 \\
    Three-Shot&\gray{87}0.69 &\gray{87}0.71 & \gray{50}0.54 & \gray{60}0.61 & \gray{50}0.5 & \gray{60}0.56 & \gray{10}0.17 & \gray{50}0.5 & \gray{10}0.09 & \gray{50}0.52 \\
    \hline
    Fine-Tuned&- &- & \gray{100}0.66 & \gray{100}0.71 & \gray{50}0.49 & \gray{60}0.57 & \gray{10}0.15 & \gray{60}0.55 & \gray{10}0.11 & \gray{50}0.48\\
    \hline 
    \end{tabular}
    \caption{\footnotesize Fine-tuning marginally improves toxicity classification accuracy in open models, but closed models still consistently outperform them.}
    \label{tab:bias_f1}
\end{table*}

\subsection{Abstention --- Fine-tuned Open Models Largely Abstain from Emitting Misinformation} 

Instruction-tuned language models answer questions based on their parametric knowledge~\cite{ouyang2022training} or based on context provided as part of the prompt by the user \cite{sanh2022multitask}. If the model has the necessary information in neither its parametric knowledge nor the user-provided context, the model should \textit{abstain} from answering to avoid misinforming a user~\cite{wen2024characterizing}.

\begin{table*}[htbp]
    \centering
    \small
    \begin{tabular}{|c|c|c|c|c|c|}
    \hline
    Scenario & GPT-4-Turbo & GPT-3.5-Turbo & Mistral-7B-Instruct & Falcon-7B-Instruct & Llama-2-7B-Chat\\ 
    \hline
    Zero-Shot&\gray{75}0.59 & \gray{68}0.54 & \gray{39}0.36 & \gray{23}0.26 & \gray{1}0.11 \\
    One-Shot&\gray{70}0.56 & \gray{55}0.46 & \gray{34}0.33 & \gray{31}0.31 & \gray{0}0.11\\
    Two-Shot&\gray{77}0.60 & \gray{35}0.34 & \gray{40}0.37  & \gray{16}0.21 & \gray{1}0.12 \\
    Three-Shot&\gray{87}0.67 & \gray{56}0.47 & \gray{46}0.41 & \gray{8}0.16 & \gray{3}0.13 \\
    \hline
    Fine-Tuned&- & \gray{100}0.74 & \gray{64}0.52 & \gray{53}0.45 & \gray{56}0.47\\
    \hline 
    \end{tabular}
    \caption{\footnotesize Fine-tuning significantly improves the performance of open models on the QASPER science question answering dataset, though open models still lag behind few-shot GPT-4-Turbo and finetuned GPT-3.5-Turbo.}
    \label{tab:abstention_f1}
\end{table*}

We evaluate the ability of open models to abstain by adapting questions from context-dependent scientific knowledge benchmarks, where some questions are designed to be unanswerable if annotators cannot find the answers based on the context.  We use the \textit{full} training set from QASPER~\cite{dasigi-etal-2021-dataset} science question answering dataset to finetune and use the \textit{answerable} questions from the test set to assess abstention in the following way: we remove the context completely, such that the correct answer is to abstain ("Without Context" in Table \ref{tab:abstention_results}). We use abstention rate to evaluate models' abstention performance following previous work~\cite{wen2024characterizing}. Ideally, the abstention rate should be 1 if we remove the context completely. In addition to abstention, we evaluate model performance on the full QASPER test set (via F1 score) to assess tradeoffs between overall performance and abstention ability (Table \ref{tab:abstention_f1}).  We follow the original split of train, validation, and test sets, resulting in 2,593, 1,005, and 1,451 questions respectively. Table~\ref{tab:abstention_f1} describes task performance on QASPER test set.  GPT-4-Turbo excels in few-shot settings. Fine-tuning significantly improves task performance across models, and fine-tuned GPT-3.5-Turbo achieves the highest F1 of 0.74, 0.07 higher than GPT-4-Turbo. Fine-tuning improves Mistral-7B-Instruct, Falcon-7B-Instruct and LLaMA-2-7B-Chat, but performance does not approach GPT-4-Turbo on this challenging task. \looseness=-1

\begin{table}[htbp]
    \centering
    \small
    \begin{tabular}{|c|c|c|}
    \hline
     \multirow{1}{*}{Scenario} & \multirow{1}{*}{Model}  & \multicolumn{1}{|c|}{Without Context} \\
    \hline
    \multirow{4}{*}{Zero-Shot} & GPT3.5-Turbo & \gray{93}0.93 \\
    & Falcon-7B-Instruct  & \gray{2}0.02  \\
    & Llama-2-7B-Chat & 0.00  \\
    & Mistral-7B-Instruct  & \gray{70}0.70  \\
    \hline
    \multirow{4}{*}{Fine-Tuned} & GPT3.5-Turbo  & \gray{53}0.53 \\
    & Falcon-7B-Instruct  & \gray{65}0.65 \\
    & Llama-2-7B-Chat  & \gray{99}0.99  \\
    & Mistral-7B-Instruct  & \gray{38}0.38 \\
    \hline
        \end{tabular}
    \caption{\footnotesize Without context, models that abstain well in the zero-shot setting  (GPT3.5 and Mistral) do not abstain well after finetuning.  Models that abstain poorly in the zero-shot setting (Falcon and Llama) \textit{improve} after finetuning.}
    \label{tab:abstention_results}
\end{table}

\noindent Table~\ref{tab:abstention_results} describes results for the abstention task ("Without Context" means the model is not provided enough information to answer the question and should always abstain) using answerable questions from QASPER test set. Surprisingly, with fine-tuning, abstention performance is reduced for the best question-answering models, suggesting an "overconfidence" effect: Models that are capable of abstaining in the zero-shot setting (GPT3.5Turbo at 0.93 and Mistral-7B-Instruct 0.70) are less likely to abstain in the fine-tuned setting (GPT3.5Turbo at 0.53 and Mistral-7B-Instruct 0.38).  However, for models that are \textit{unable} to abstain in the zero-shot setting (Falcon-7B-Instruct at 0.02 and Llama-2-7B-Chat at 0.00), fine-tuning significantly improves this capability (Falcon-7B-Instruct at 0.65 and Llama-2-7B-Chat at 0.99).  Results suggest a sweet spot in balancing overall performance with the ability to abstain using ordinary training regimes.

\looseness=-1

\section{Discussion}

\subsection{The Viability and Implications of Laboratory-Scale AI.} Our work provides empirical support for the viability of adopting a ``laboratory-scale'' approach to AI that prioritizes user autonomy, privacy, fairness, and transparency while maintaining much of the performance and usability offered by industry-dominant corporate models. With a small GPU card, users can create domain-specific, chat-based language models and deploy them without losing the general-purpose utility and interface that makes such technologies appealing. The laboratory-scale approach intends to address, in a limited capacity, the challenges posed by scholars such as \citet{bender2021dangers}, who highlight the dangers of training language models on poorly specified web scraped data generally unrelated to the tasks for which the model will be used; \citet{birhane2022values}, who describe the performance-centric ''values encoded in machine learning research,'' and highlight the field's capture by big tech companies; and \citet{palmer2023using}, who contend that scientists and academic researchers must justify the use of proprietary, closed models over open models. Laboratory-scale AI centers the domain-specific, responsible application of small, open models, presenting an option for scientists and public interest technologists who have good reason to avoid closed models that cannot be accessed except via a call to an API. \looseness=-1

\subsection{Affordances and Challenges of Open Models.} We used the libraries and model ecosystem provided by \href{https://huggingface.co/}{HuggingFace} \cite{wolf2019huggingface}. The Supervised Fine-Tuning trainer class provided by the TRL library made adapting open language models relatively simple, and primarily dependent on the organization of our data. The Huggingface ecosystem also supports qLoRA \cite{dettmers2023qlora}, which made customizing quantized models relatively straightforward. However, we nonetheless encountered difficulties with using open models that bear discussion. The most intractable problem we encountered in fine-tuning our own models lay in the difficulty of obtaining cloud instances equipped with even low-cost GPU hardware. We experienced consistent difficulties obtaining results due to lack of available cloud resources. Moreover, we did not expect the quantized open models we tested to run so much more slowly than the closed models we tested. This is related in part to our choice of a low-end GPU, but where inference speed makes a difference, the evidence suggests that cost-efficient, laboratory-scale models still trail closed models. \looseness=-1

Open models showed room for improvement on tasks related to responsible use and deployment. While results on differentially private question answering show the potential for privacy-centering open models, they are impeded by small batch sizes required to use low-cost hardware. Fine-tuning has mixed effects for abstention: where a model exhibits strong question answering performance, it is less likely to abstain when it should; but when it exhibits weak question answering performance, it more reliably abstains from answering a question when it should not. Results on the toxicity bias task suggest that open models lag behind closed models on bias mitigation. Though tempting to conclude that the RLHF process used by OpenAI is the right way to address this problem, we note that LLaMA-2-Chat-7B also undergoes RLHF \cite{touvron2023llama2}, and performs most poorly of any of the models we assess. Future research can contribute by centering these issues. \looseness=-1

\subsection{Limitations and Future Work.} While our work attempts to provide an open, low-cost approach, we acknowledge that open models have undergone expensive, resource intensive pretraining on large-scale, sometimes opaque datasets. While libraries like qLoRA help to enable adaptations of pretrained models, they cannot equip us with a means of circumventing pretraining, which at this time remains the only reliable means of producing a fluent, general-purpose base model. Future work might explore alternatives that change the pretraining paradigm. We also acknowledge that results from closed models may not be reproducible, should OpenAI change or remove models from its API, potentially without notifying the end user. This is a limitation of closed models that motivates our study, but also necessarily a limitation of our work. Finally, we could not reliably model the carbon cost of closed models due to uncertainties about the exact hardware used to run these models, the location of the data centers on which they run, and practices such as batching user inputs, which may allow for economies of scale. \looseness=-1

\section{Conclusion}

We show that small, open, models are competitive with closed models, in that they are cost-efficient, responsive to user data, and robust to fine-tuning. We analyze responsible use of lab-scale models, showing they offer benefits over closed models, particularly in privacy and abstention. We contend that laboratory-scale AI can serve as a basis for future scientific and public interest work, enabling practitioners to customize models without needing to rely on closed, API-based AI. \looseness=-1

\bibliographystyle{ACM-Reference-Format}
\bibliography{references}


\begin{thebibliography}{77}


\ifx \showCODEN    \undefined \def \showCODEN     #1{\unskip}     \fi
\ifx \showDOI      \undefined \def \showDOI       #1{#1}\fi
\ifx \showISBNx    \undefined \def \showISBNx     #1{\unskip}     \fi
\ifx \showISBNxiii \undefined \def \showISBNxiii  #1{\unskip}     \fi
\ifx \showISSN     \undefined \def \showISSN      #1{\unskip}     \fi
\ifx \showLCCN     \undefined \def \showLCCN      #1{\unskip}     \fi
\ifx \shownote     \undefined \def \shownote      #1{#1}          \fi
\ifx \showarticletitle \undefined \def \showarticletitle #1{#1}   \fi
\ifx \showURL      \undefined \def \showURL       {\relax}        \fi
\providecommand\bibfield[2]{#2}
\providecommand\bibinfo[2]{#2}
\providecommand\natexlab[1]{#1}
\providecommand\showeprint[2][]{arXiv:#2}

\bibitem[Abadi et~al\mbox{.}(2016)]%
        {abadi2016deep}
\bibfield{author}{\bibinfo{person}{Martin Abadi}, \bibinfo{person}{Andy Chu}, \bibinfo{person}{Ian Goodfellow}, \bibinfo{person}{H~Brendan McMahan}, \bibinfo{person}{Ilya Mironov}, \bibinfo{person}{Kunal Talwar}, {and} \bibinfo{person}{Li Zhang}.} \bibinfo{year}{2016}\natexlab{}.
\newblock \showarticletitle{Deep learning with differential privacy}. In \bibinfo{booktitle}{\emph{Proceedings of the 2016 ACM SIGSAC conference on computer and communications security}}. \bibinfo{publisher}{{}}, \bibinfo{address}{{}}, \bibinfo{pages}{308--318}.
\newblock


\bibitem[Almazrouei et~al\mbox{.}(2023)]%
        {almazrouei2023falcon}
\bibfield{author}{\bibinfo{person}{Ebtesam Almazrouei}, \bibinfo{person}{Hamza Alobeidli}, \bibinfo{person}{Abdulaziz Alshamsi}, \bibinfo{person}{Alessandro Cappelli}, \bibinfo{person}{Ruxandra Cojocaru}, \bibinfo{person}{M{\'e}rouane Debbah}, \bibinfo{person}{{\'E}tienne Goffinet}, \bibinfo{person}{Daniel Hesslow}, \bibinfo{person}{Julien Launay}, \bibinfo{person}{Quentin Malartic}, {et~al\mbox{.}}} \bibinfo{year}{2023}\natexlab{}.
\newblock \showarticletitle{The Falcon Series of Open Language Models}.
\newblock \bibinfo{journal}{\emph{arXiv preprint arXiv:2311.16867}} \bibinfo{volume}{{}}, \bibinfo{number}{{}} (\bibinfo{year}{2023}), \bibinfo{pages}{{}}.
\newblock


\bibitem[Bai et~al\mbox{.}(2022)]%
        {bai2022training}
\bibfield{author}{\bibinfo{person}{Yuntao Bai}, \bibinfo{person}{Andy Jones}, \bibinfo{person}{Kamal Ndousse}, \bibinfo{person}{Amanda Askell}, \bibinfo{person}{Anna Chen}, \bibinfo{person}{Nova DasSarma}, \bibinfo{person}{Dawn Drain}, \bibinfo{person}{Stanislav Fort}, \bibinfo{person}{Deep Ganguli}, \bibinfo{person}{Tom Henighan}, {et~al\mbox{.}}} \bibinfo{year}{2022}\natexlab{}.
\newblock \showarticletitle{Training a helpful and harmless assistant with reinforcement learning from human feedback}.
\newblock \bibinfo{journal}{\emph{arXiv preprint arXiv:2204.05862}} \bibinfo{volume}{{}}, \bibinfo{number}{{}} (\bibinfo{year}{2022}), \bibinfo{pages}{{}}.
\newblock


\bibitem[Beckman(2023)]%
        {openaistats}
\bibfield{author}{\bibinfo{person}{Jeff Beckman}.} \bibinfo{year}{2023}\natexlab{}.
\newblock \bibinfo{title}{OpenAI Statistics 2023: Growth, Users, and More}.
\newblock \bibinfo{howpublished}{\url{https://techreport.com/statistics/openai-statistics/}}.
\newblock
\newblock
\shownote{[Accessed 19-01-2024]}.


\bibitem[Ben~Abacha et~al\mbox{.}(2023)]%
        {mts-dialog}
\bibfield{author}{\bibinfo{person}{Asma Ben~Abacha}, \bibinfo{person}{Wen-wai Yim}, \bibinfo{person}{Yadan Fan}, {and} \bibinfo{person}{Thomas Lin}.} \bibinfo{year}{2023}\natexlab{}.
\newblock \showarticletitle{An Empirical Study of Clinical Note Generation from Doctor-Patient Encounters}. In \bibinfo{booktitle}{\emph{Proceedings of the 17th Conference of the European Chapter of the Association for Computational Linguistics}}. \bibinfo{publisher}{Association for Computational Linguistics}, \bibinfo{address}{Dubrovnik, Croatia}, \bibinfo{pages}{2291--2302}.
\newblock
\urldef\tempurl%
\url{https://aclanthology.org/2023.eacl-main.168}
\showURL{%
\tempurl}


\bibitem[Bender et~al\mbox{.}(2021)]%
        {bender2021dangers}
\bibfield{author}{\bibinfo{person}{Emily~M Bender}, \bibinfo{person}{Timnit Gebru}, \bibinfo{person}{Angelina McMillan-Major}, {and} \bibinfo{person}{Shmargaret Shmitchell}.} \bibinfo{year}{2021}\natexlab{}.
\newblock \showarticletitle{On the dangers of stochastic parrots: Can language models be too big?}. In \bibinfo{booktitle}{\emph{Proceedings of the 2021 ACM conference on fairness, accountability, and transparency}}. \bibinfo{publisher}{{}}, \bibinfo{address}{{}}, \bibinfo{pages}{610--623}.
\newblock


\bibitem[Biewald(2020)]%
        {wandb}
\bibfield{author}{\bibinfo{person}{Lukas Biewald}.} \bibinfo{year}{2020}\natexlab{}.
\newblock \bibinfo{title}{Experiment Tracking with Weights and Biases}.
\newblock
\newblock
\urldef\tempurl%
\url{https://www.wandb.com/}
\showURL{%
\tempurl}
\newblock
\shownote{Software available from wandb.com}.


\bibitem[Birhane et~al\mbox{.}(2022)]%
        {birhane2022values}
\bibfield{author}{\bibinfo{person}{Abeba Birhane}, \bibinfo{person}{Pratyusha Kalluri}, \bibinfo{person}{Dallas Card}, \bibinfo{person}{William Agnew}, \bibinfo{person}{Ravit Dotan}, {and} \bibinfo{person}{Michelle Bao}.} \bibinfo{year}{2022}\natexlab{}.
\newblock \showarticletitle{The values encoded in machine learning research}. In \bibinfo{booktitle}{\emph{Proceedings of the 2022 ACM Conference on Fairness, Accountability, and Transparency}}. \bibinfo{publisher}{{}}, \bibinfo{address}{{}}, \bibinfo{pages}{173--184}.
\newblock


\bibitem[Bisong and Bisong(2019)]%
        {bisong2019overview}
\bibfield{author}{\bibinfo{person}{Ekaba Bisong} {and} \bibinfo{person}{Ekaba Bisong}.} \bibinfo{year}{2019}\natexlab{}.
\newblock \showarticletitle{An overview of google cloud platform services}.
\newblock \bibinfo{journal}{\emph{Building Machine Learning and Deep Learning Models on Google Cloud Platform: A Comprehensive Guide for Beginners}} \bibinfo{volume}{{}}, \bibinfo{number}{{}} (\bibinfo{year}{2019}), \bibinfo{pages}{7--10}.
\newblock


\bibitem[Bommasani et~al\mbox{.}(2023)]%
        {bommasani2023foundation}
\bibfield{author}{\bibinfo{person}{Rishi Bommasani}, \bibinfo{person}{Kevin Klyman}, \bibinfo{person}{Shayne Longpre}, \bibinfo{person}{Sayash Kapoor}, \bibinfo{person}{Nestor Maslej}, \bibinfo{person}{Betty Xiong}, \bibinfo{person}{Daniel Zhang}, {and} \bibinfo{person}{Percy Liang}.} \bibinfo{year}{2023}\natexlab{}.
\newblock \showarticletitle{The foundation model transparency index}.
\newblock \bibinfo{journal}{\emph{arXiv preprint arXiv:2310.12941}} \bibinfo{volume}{{}}, \bibinfo{number}{{}} (\bibinfo{year}{2023}), \bibinfo{pages}{{}}.
\newblock


\bibitem[Brown et~al\mbox{.}(2020)]%
        {brown2020language}
\bibfield{author}{\bibinfo{person}{Tom Brown}, \bibinfo{person}{Benjamin Mann}, \bibinfo{person}{Nick Ryder}, \bibinfo{person}{Melanie Subbiah}, \bibinfo{person}{Jared~D Kaplan}, \bibinfo{person}{Prafulla Dhariwal}, \bibinfo{person}{Arvind Neelakantan}, \bibinfo{person}{Pranav Shyam}, \bibinfo{person}{Girish Sastry}, \bibinfo{person}{Amanda Askell}, {et~al\mbox{.}}} \bibinfo{year}{2020}\natexlab{}.
\newblock \showarticletitle{Language models are few-shot learners}.
\newblock \bibinfo{journal}{\emph{Advances in neural information processing systems}}  \bibinfo{volume}{33} (\bibinfo{year}{2020}), \bibinfo{pages}{1877--1901}.
\newblock


\bibitem[Bulatov(2018)]%
        {gradientcheckpointing}
\bibfield{author}{\bibinfo{person}{Yaroslav Bulatov}.} \bibinfo{year}{2018}\natexlab{}.
\newblock \bibinfo{title}{Fitting larger networks into memory.}
\newblock \bibinfo{howpublished}{\url{https://medium.com/tensorflow/fitting-larger-networks-into-memory-583e3c758ff9}}.
\newblock
\newblock
\shownote{[Accessed 19-01-2024]}.


\bibitem[Chalkidis et~al\mbox{.}(2020)]%
        {chalkidis2020legal}
\bibfield{author}{\bibinfo{person}{Ilias Chalkidis}, \bibinfo{person}{Manos Fergadiotis}, \bibinfo{person}{Prodromos Malakasiotis}, \bibinfo{person}{Nikolaos Aletras}, {and} \bibinfo{person}{Ion Androutsopoulos}.} \bibinfo{year}{2020}\natexlab{}.
\newblock \showarticletitle{LEGAL-BERT: The Muppets straight out of Law School}. In \bibinfo{booktitle}{\emph{Findings of the Association for Computational Linguistics: EMNLP 2020}}. \bibinfo{publisher}{{}}, \bibinfo{address}{{}}, \bibinfo{pages}{2898--2904}.
\newblock


\bibitem[Chee et~al\mbox{.}(2023)]%
        {chee2023quip}
\bibfield{author}{\bibinfo{person}{Jerry Chee}, \bibinfo{person}{Yaohui Cai}, \bibinfo{person}{Volodymyr Kuleshov}, {and} \bibinfo{person}{Christopher De~Sa}.} \bibinfo{year}{2023}\natexlab{}.
\newblock \showarticletitle{Quip: 2-bit quantization of large language models with guarantees}.
\newblock \bibinfo{journal}{\emph{arXiv preprint arXiv:2307.13304}} \bibinfo{volume}{{}}, \bibinfo{number}{{}} (\bibinfo{year}{2023}), \bibinfo{pages}{{}}.
\newblock


\bibitem[Chen et~al\mbox{.}(2016)]%
        {chen2016training}
\bibfield{author}{\bibinfo{person}{Tianqi Chen}, \bibinfo{person}{Bing Xu}, \bibinfo{person}{Chiyuan Zhang}, {and} \bibinfo{person}{Carlos Guestrin}.} \bibinfo{year}{2016}\natexlab{}.
\newblock \showarticletitle{Training deep nets with sublinear memory cost}.
\newblock \bibinfo{journal}{\emph{arXiv preprint arXiv:1604.06174}} \bibinfo{volume}{{}}, \bibinfo{number}{{}} (\bibinfo{year}{2016}), \bibinfo{pages}{{}}.
\newblock


\bibitem[Colombo et~al\mbox{.}(2022)]%
        {colombo2022best}
\bibfield{author}{\bibinfo{person}{Pierre Colombo}, \bibinfo{person}{Nathan Noiry}, \bibinfo{person}{Ekhine Irurozki}, {and} \bibinfo{person}{Stephan Clemencon}.} \bibinfo{year}{2022}\natexlab{}.
\newblock \showarticletitle{What are the best systems? New perspectives on NLP Benchmarking}.
\newblock \bibinfo{journal}{\emph{arXiv preprint arXiv:2202.03799}} \bibinfo{volume}{{}}, \bibinfo{number}{{}} (\bibinfo{year}{2022}), \bibinfo{pages}{{}}.
\newblock


\bibitem[Dai and Le(2015)]%
        {dai2015semi}
\bibfield{author}{\bibinfo{person}{Andrew~M Dai} {and} \bibinfo{person}{Quoc~V Le}.} \bibinfo{year}{2015}\natexlab{}.
\newblock \showarticletitle{Semi-supervised sequence learning}.
\newblock \bibinfo{journal}{\emph{Advances in neural information processing systems}} \bibinfo{volume}{28}, \bibinfo{number}{{}} (\bibinfo{year}{2015}), \bibinfo{pages}{{}}.
\newblock


\bibitem[Dang et~al\mbox{.}(2022)]%
        {dang2022prompt}
\bibfield{author}{\bibinfo{person}{Hai Dang}, \bibinfo{person}{Lukas Mecke}, \bibinfo{person}{Florian Lehmann}, \bibinfo{person}{Sven Goller}, {and} \bibinfo{person}{Daniel Buschek}.} \bibinfo{year}{2022}\natexlab{}.
\newblock \showarticletitle{How to prompt? Opportunities and challenges of zero-and few-shot learning for human-AI interaction in creative applications of generative models}.
\newblock \bibinfo{journal}{\emph{arXiv preprint arXiv:2209.01390}} \bibinfo{volume}{{}}, \bibinfo{number}{{}} (\bibinfo{year}{2022}), \bibinfo{pages}{{}}.
\newblock


\bibitem[Dasigi et~al\mbox{.}(2021)]%
        {dasigi-etal-2021-dataset}
\bibfield{author}{\bibinfo{person}{Pradeep Dasigi}, \bibinfo{person}{Kyle Lo}, \bibinfo{person}{Iz Beltagy}, \bibinfo{person}{Arman Cohan}, \bibinfo{person}{Noah~A. Smith}, {and} \bibinfo{person}{Matt Gardner}.} \bibinfo{year}{2021}\natexlab{}.
\newblock \showarticletitle{A Dataset of Information-Seeking Questions and Answers Anchored in Research Papers}. In \bibinfo{booktitle}{\emph{Proceedings of the 2021 Conference of the North American Chapter of the Association for Computational Linguistics: Human Language Technologies}}, \bibfield{editor}{\bibinfo{person}{Kristina Toutanova}, \bibinfo{person}{Anna Rumshisky}, \bibinfo{person}{Luke Zettlemoyer}, \bibinfo{person}{Dilek Hakkani-Tur}, \bibinfo{person}{Iz~Beltagy}, \bibinfo{person}{Steven Bethard}, \bibinfo{person}{Ryan Cotterell}, \bibinfo{person}{Tanmoy Chakraborty}, {and} \bibinfo{person}{Yichao Zhou}} (Eds.). \bibinfo{publisher}{Association for Computational Linguistics}, \bibinfo{address}{Online}, \bibinfo{pages}{4599--4610}.
\newblock
\urldef\tempurl%
\url{https://doi.org/10.18653/v1/2021.naacl-main.365}
\showDOI{\tempurl}


\bibitem[Dettmers et~al\mbox{.}(2022)]%
        {dettmers2022llm}
\bibfield{author}{\bibinfo{person}{Tim Dettmers}, \bibinfo{person}{Mike Lewis}, \bibinfo{person}{Younes Belkada}, {and} \bibinfo{person}{Luke Zettlemoyer}.} \bibinfo{year}{2022}\natexlab{}.
\newblock \showarticletitle{Llm. int8 (): 8-bit matrix multiplication for transformers at scale}.
\newblock \bibinfo{journal}{\emph{arXiv preprint arXiv:2208.07339}} \bibinfo{volume}{{}}, \bibinfo{number}{{}} (\bibinfo{year}{2022}), \bibinfo{pages}{{}}.
\newblock


\bibitem[Dettmers et~al\mbox{.}(2023)]%
        {dettmers2023qlora}
\bibfield{author}{\bibinfo{person}{Tim Dettmers}, \bibinfo{person}{Artidoro Pagnoni}, \bibinfo{person}{Ari Holtzman}, {and} \bibinfo{person}{Luke Zettlemoyer}.} \bibinfo{year}{2023}\natexlab{}.
\newblock \showarticletitle{Qlora: Efficient finetuning of quantized llms}.
\newblock \bibinfo{journal}{\emph{arXiv preprint arXiv:2305.14314}} \bibinfo{volume}{{}}, \bibinfo{number}{{}} (\bibinfo{year}{2023}), \bibinfo{pages}{{}}.
\newblock


\bibitem[Dettmers and Zettlemoyer(2023)]%
        {dettmers2023case}
\bibfield{author}{\bibinfo{person}{Tim Dettmers} {and} \bibinfo{person}{Luke Zettlemoyer}.} \bibinfo{year}{2023}\natexlab{}.
\newblock \showarticletitle{The case for 4-bit precision: k-bit inference scaling laws}. In \bibinfo{booktitle}{\emph{International Conference on Machine Learning}}. PMLR, \bibinfo{publisher}{{}}, \bibinfo{address}{{}}, \bibinfo{pages}{7750--7774}.
\newblock


\bibitem[Diggelmann et~al\mbox{.}(2020)]%
        {diggelmann2020climate}
\bibfield{author}{\bibinfo{person}{Thomas Diggelmann}, \bibinfo{person}{Jordan Boyd-Graber}, \bibinfo{person}{Jannis Bulian}, \bibinfo{person}{Massimiliano Ciaramita}, {and} \bibinfo{person}{Markus Leippold}.} \bibinfo{year}{2020}\natexlab{}.
\newblock \showarticletitle{Climate-fever: A dataset for verification of real-world climate claims}.
\newblock \bibinfo{journal}{\emph{arXiv preprint arXiv:2012.00614}} \bibinfo{volume}{{}}, \bibinfo{number}{{}} (\bibinfo{year}{2020}), \bibinfo{pages}{{}}.
\newblock


\bibitem[Ding et~al\mbox{.}(2023)]%
        {ding2023parameter}
\bibfield{author}{\bibinfo{person}{Ning Ding}, \bibinfo{person}{Yujia Qin}, \bibinfo{person}{Guang Yang}, \bibinfo{person}{Fuchao Wei}, \bibinfo{person}{Zonghan Yang}, \bibinfo{person}{Yusheng Su}, \bibinfo{person}{Shengding Hu}, \bibinfo{person}{Yulin Chen}, \bibinfo{person}{Chi-Min Chan}, \bibinfo{person}{Weize Chen}, {et~al\mbox{.}}} \bibinfo{year}{2023}\natexlab{}.
\newblock \showarticletitle{Parameter-efficient fine-tuning of large-scale pre-trained language models}.
\newblock \bibinfo{journal}{\emph{Nature Machine Intelligence}} \bibinfo{volume}{5}, \bibinfo{number}{3} (\bibinfo{year}{2023}), \bibinfo{pages}{220--235}.
\newblock


\bibitem[Fatos(2024)]%
        {aosfatos}
\bibfield{author}{\bibinfo{person}{Aos Fatos}.} \bibinfo{year}{2024}\natexlab{}.
\newblock \bibinfo{title}{Aos Fatos}.
\newblock \bibinfo{howpublished}{\url{https://www.aosfatos.org/}}.
\newblock
\newblock
\shownote{[Accessed 22-01-2024]}.


\bibitem[Getoor and Machanavajjhala(2012)]%
        {10.14778/2367502.2367564}
\bibfield{author}{\bibinfo{person}{Lise Getoor} {and} \bibinfo{person}{Ashwin Machanavajjhala}.} \bibinfo{year}{2012}\natexlab{}.
\newblock \showarticletitle{Entity resolution: theory, practice \& open challenges}.
\newblock \bibinfo{journal}{\emph{Proc. VLDB Endow.}} \bibinfo{volume}{5}, \bibinfo{number}{12} (\bibinfo{date}{aug} \bibinfo{year}{2012}), \bibinfo{pages}{2018–2019}.
\newblock
\showISSN{2150-8097}
\urldef\tempurl%
\url{https://doi.org/10.14778/2367502.2367564}
\showDOI{\tempurl}


\bibitem[Ghazi et~al\mbox{.}(2021)]%
        {ghazi2021deep}
\bibfield{author}{\bibinfo{person}{Badih Ghazi}, \bibinfo{person}{Noah Golowich}, \bibinfo{person}{Ravi Kumar}, \bibinfo{person}{Pasin Manurangsi}, {and} \bibinfo{person}{Chiyuan Zhang}.} \bibinfo{year}{2021}\natexlab{}.
\newblock \showarticletitle{Deep learning with label differential privacy}.
\newblock \bibinfo{journal}{\emph{Advances in neural information processing systems}}  \bibinfo{volume}{34} (\bibinfo{year}{2021}), \bibinfo{pages}{27131--27145}.
\newblock


\bibitem[Gilardi et~al\mbox{.}(2023)]%
        {gilardi2023chatgpt}
\bibfield{author}{\bibinfo{person}{Fabrizio Gilardi}, \bibinfo{person}{Meysam Alizadeh}, {and} \bibinfo{person}{Ma{\"e}l Kubli}.} \bibinfo{year}{2023}\natexlab{}.
\newblock \showarticletitle{ChatGPT outperforms crowd workers for text-annotation tasks}.
\newblock \bibinfo{journal}{\emph{Proceedings of the National Academy of Sciences}} \bibinfo{volume}{120}, \bibinfo{number}{30} (\bibinfo{year}{2023}), \bibinfo{pages}{e2305016120}.
\newblock


\bibitem[Gong et~al\mbox{.}(2020)]%
        {gong2020survey}
\bibfield{author}{\bibinfo{person}{Maoguo Gong}, \bibinfo{person}{Yu Xie}, \bibinfo{person}{Ke Pan}, \bibinfo{person}{Kaiyuan Feng}, {and} \bibinfo{person}{Alex~Kai Qin}.} \bibinfo{year}{2020}\natexlab{}.
\newblock \showarticletitle{A survey on differentially private machine learning}.
\newblock \bibinfo{journal}{\emph{IEEE computational intelligence magazine}} \bibinfo{volume}{15}, \bibinfo{number}{2} (\bibinfo{year}{2020}), \bibinfo{pages}{49--64}.
\newblock


\bibitem[Han et~al\mbox{.}(2023)]%
        {han2023huskyscribe}
\bibfield{author}{\bibinfo{person}{Bin Han}, \bibinfo{person}{Haotian Zhu}, \bibinfo{person}{Sitong Zhou}, \bibinfo{person}{Sofia Ahmed}, \bibinfo{person}{M Rahman}, \bibinfo{person}{Fei Xia}, {and} \bibinfo{person}{Kevin Lybarger}.} \bibinfo{year}{2023}\natexlab{}.
\newblock \showarticletitle{Huskyscribe at mediqa-sum 2023: Summarizing clinical dialogues with transformers}. In \bibinfo{booktitle}{\emph{{}}}. CLEF, \bibinfo{publisher}{{}}, \bibinfo{address}{{}}, \bibinfo{pages}{{}}.
\newblock


\bibitem[Hendrycks et~al\mbox{.}(2021)]%
        {hendrycks2021measuring}
\bibfield{author}{\bibinfo{person}{Dan Hendrycks}, \bibinfo{person}{Collin Burns}, \bibinfo{person}{Steven Basart}, \bibinfo{person}{Andy Zou}, \bibinfo{person}{Mantas Mazeika}, \bibinfo{person}{Dawn Song}, {and} \bibinfo{person}{Jacob Steinhardt}.} \bibinfo{year}{2021}\natexlab{}.
\newblock \bibinfo{title}{Measuring Massive Multitask Language Understanding}.
\newblock
\newblock
\showeprint[arxiv]{2009.03300}~[cs.CY]


\bibitem[Hu et~al\mbox{.}(2021)]%
        {hu2021lora}
\bibfield{author}{\bibinfo{person}{Edward~J Hu}, \bibinfo{person}{Yelong Shen}, \bibinfo{person}{Phillip Wallis}, \bibinfo{person}{Zeyuan Allen-Zhu}, \bibinfo{person}{Yuanzhi Li}, \bibinfo{person}{Shean Wang}, \bibinfo{person}{Lu Wang}, {and} \bibinfo{person}{Weizhu Chen}.} \bibinfo{year}{2021}\natexlab{}.
\newblock \showarticletitle{Lora: Low-rank adaptation of large language models}.
\newblock \bibinfo{journal}{\emph{arXiv preprint arXiv:2106.09685}} \bibinfo{volume}{{}}, \bibinfo{number}{{}} (\bibinfo{year}{2021}), \bibinfo{pages}{{}}.
\newblock


\bibitem[Jiang et~al\mbox{.}(2023)]%
        {jiang2023mistral}
\bibfield{author}{\bibinfo{person}{Albert~Q Jiang}, \bibinfo{person}{Alexandre Sablayrolles}, \bibinfo{person}{Arthur Mensch}, \bibinfo{person}{Chris Bamford}, \bibinfo{person}{Devendra~Singh Chaplot}, \bibinfo{person}{Diego de~las Casas}, \bibinfo{person}{Florian Bressand}, \bibinfo{person}{Gianna Lengyel}, \bibinfo{person}{Guillaume Lample}, \bibinfo{person}{Lucile Saulnier}, {et~al\mbox{.}}} \bibinfo{year}{2023}\natexlab{}.
\newblock \showarticletitle{Mistral 7B}.
\newblock \bibinfo{journal}{\emph{arXiv preprint arXiv:2310.06825}} \bibinfo{volume}{{}}, \bibinfo{number}{{}} (\bibinfo{year}{2023}), \bibinfo{pages}{{}}.
\newblock


\bibitem[Jin et~al\mbox{.}(2021)]%
        {jin2021disease}
\bibfield{author}{\bibinfo{person}{Di Jin}, \bibinfo{person}{Eileen Pan}, \bibinfo{person}{Nassim Oufattole}, \bibinfo{person}{Wei-Hung Weng}, \bibinfo{person}{Hanyi Fang}, {and} \bibinfo{person}{Peter Szolovits}.} \bibinfo{year}{2021}\natexlab{}.
\newblock \showarticletitle{What disease does this patient have? a large-scale open domain question answering dataset from medical exams}.
\newblock \bibinfo{journal}{\emph{Applied Sciences}} \bibinfo{volume}{11}, \bibinfo{number}{14} (\bibinfo{year}{2021}), \bibinfo{pages}{6421}.
\newblock


\bibitem[Koco{\'n} et~al\mbox{.}(2023)]%
        {kocon2023chatgpt}
\bibfield{author}{\bibinfo{person}{Jan Koco{\'n}}, \bibinfo{person}{Igor Cichecki}, \bibinfo{person}{Oliwier Kaszyca}, \bibinfo{person}{Mateusz Kochanek}, \bibinfo{person}{Dominika Szyd{\l}o}, \bibinfo{person}{Joanna Baran}, \bibinfo{person}{Julita Bielaniewicz}, \bibinfo{person}{Marcin Gruza}, \bibinfo{person}{Arkadiusz Janz}, \bibinfo{person}{Kamil Kanclerz}, {et~al\mbox{.}}} \bibinfo{year}{2023}\natexlab{}.
\newblock \showarticletitle{ChatGPT: Jack of all trades, master of none}.
\newblock \bibinfo{journal}{\emph{Information Fusion}} \bibinfo{volume}{{}}, \bibinfo{number}{{}} (\bibinfo{year}{2023}), \bibinfo{pages}{101861}.
\newblock


\bibitem[Koh et~al\mbox{.}(2021)]%
        {koh2021wilds}
\bibfield{author}{\bibinfo{person}{Pang~Wei Koh}, \bibinfo{person}{Shiori Sagawa}, \bibinfo{person}{Henrik Marklund}, \bibinfo{person}{Sang~Michael Xie}, \bibinfo{person}{Marvin Zhang}, \bibinfo{person}{Akshay Balsubramani}, \bibinfo{person}{Weihua Hu}, \bibinfo{person}{Michihiro Yasunaga}, \bibinfo{person}{Richard~Lanas Phillips}, \bibinfo{person}{Irena Gao}, {et~al\mbox{.}}} \bibinfo{year}{2021}\natexlab{}.
\newblock \showarticletitle{Wilds: A benchmark of in-the-wild distribution shifts}. In \bibinfo{booktitle}{\emph{International Conference on Machine Learning}}. PMLR, \bibinfo{publisher}{{}}, \bibinfo{address}{{}}, \bibinfo{pages}{5637--5664}.
\newblock


\bibitem[Liang et~al\mbox{.}(2023)]%
        {liang2023holistic}
\bibfield{author}{\bibinfo{person}{Percy Liang}, \bibinfo{person}{Rishi Bommasani}, \bibinfo{person}{Tony Lee}, \bibinfo{person}{Dimitris Tsipras}, \bibinfo{person}{Dilara Soylu}, \bibinfo{person}{Michihiro Yasunaga}, \bibinfo{person}{Yian Zhang}, \bibinfo{person}{Deepak Narayanan}, \bibinfo{person}{Yuhuai Wu}, \bibinfo{person}{Ananya Kumar}, \bibinfo{person}{Benjamin Newman}, \bibinfo{person}{Binhang Yuan}, \bibinfo{person}{Bobby Yan}, \bibinfo{person}{Ce Zhang}, \bibinfo{person}{Christian Cosgrove}, \bibinfo{person}{Christopher~D. Manning}, \bibinfo{person}{Christopher Ré}, \bibinfo{person}{Diana Acosta-Navas}, \bibinfo{person}{Drew~A. Hudson}, \bibinfo{person}{Eric Zelikman}, \bibinfo{person}{Esin Durmus}, \bibinfo{person}{Faisal Ladhak}, \bibinfo{person}{Frieda Rong}, \bibinfo{person}{Hongyu Ren}, \bibinfo{person}{Huaxiu Yao}, \bibinfo{person}{Jue Wang}, \bibinfo{person}{Keshav Santhanam}, \bibinfo{person}{Laurel Orr}, \bibinfo{person}{Lucia Zheng}, \bibinfo{person}{Mert Yuksekgonul},
  \bibinfo{person}{Mirac Suzgun}, \bibinfo{person}{Nathan Kim}, \bibinfo{person}{Neel Guha}, \bibinfo{person}{Niladri Chatterji}, \bibinfo{person}{Omar Khattab}, \bibinfo{person}{Peter Henderson}, \bibinfo{person}{Qian Huang}, \bibinfo{person}{Ryan Chi}, \bibinfo{person}{Sang~Michael Xie}, \bibinfo{person}{Shibani Santurkar}, \bibinfo{person}{Surya Ganguli}, \bibinfo{person}{Tatsunori Hashimoto}, \bibinfo{person}{Thomas Icard}, \bibinfo{person}{Tianyi Zhang}, \bibinfo{person}{Vishrav Chaudhary}, \bibinfo{person}{William Wang}, \bibinfo{person}{Xuechen Li}, \bibinfo{person}{Yifan Mai}, \bibinfo{person}{Yuhui Zhang}, {and} \bibinfo{person}{Yuta Koreeda}.} \bibinfo{year}{2023}\natexlab{}.
\newblock \bibinfo{title}{Holistic Evaluation of Language Models}.
\newblock
\newblock
\showeprint[arxiv]{2211.09110}~[cs.CL]


\bibitem[Liesenfeld et~al\mbox{.}(2023)]%
        {liesenfeld2023opening}
\bibfield{author}{\bibinfo{person}{Andreas Liesenfeld}, \bibinfo{person}{Alianda Lopez}, {and} \bibinfo{person}{Mark Dingemanse}.} \bibinfo{year}{2023}\natexlab{}.
\newblock \showarticletitle{Opening up ChatGPT: Tracking openness, transparency, and accountability in instruction-tuned text generators}. In \bibinfo{booktitle}{\emph{Proceedings of the 5th international conference on conversational user interfaces}}. \bibinfo{publisher}{{}}, \bibinfo{address}{{}}, \bibinfo{pages}{1--6}.
\newblock


\bibitem[Loukas et~al\mbox{.}(2023)]%
        {liang-etal-2023-breaking}
\bibfield{author}{\bibinfo{person}{Lefteris Loukas}, \bibinfo{person}{Ilias Stogiannidis}, \bibinfo{person}{Prodromos Malakasiotis}, {and} \bibinfo{person}{Stavros Vassos}.} \bibinfo{year}{2023}\natexlab{}.
\newblock \showarticletitle{Breaking the Bank with {C}hat{GPT}: Few-Shot Text Classification for Finance}. In \bibinfo{booktitle}{\emph{Proceedings of the Fifth Workshop on Financial Technology and Natural Language Processing and the Second Multimodal AI For Financial Forecasting}}, \bibfield{editor}{\bibinfo{person}{Chung-Chi Chen}, \bibinfo{person}{Hiroya Takamura}, \bibinfo{person}{Puneet Mathur}, \bibinfo{person}{Remit Sawhney}, \bibinfo{person}{Hen-Hsen Huang}, {and} \bibinfo{person}{Hsin-Hsi Chen}} (Eds.). \bibinfo{publisher}{-}, \bibinfo{address}{Macao}, \bibinfo{pages}{74--80}.
\newblock
\urldef\tempurl%
\url{https://aclanthology.org/2023.finnlp-1.7}
\showURL{%
\tempurl}


\bibitem[Meedan(2024)]%
        {meedan}
\bibfield{author}{\bibinfo{person}{Meedan}.} \bibinfo{year}{2024}\natexlab{}.
\newblock \bibinfo{title}{Meedan}.
\newblock \bibinfo{howpublished}{\url{https://meedan.com/}}.
\newblock
\newblock
\shownote{[Accessed 22-01-2024]}.


\bibitem[OpenAI et~al\mbox{.}(2023)]%
        {openai2023gpt4}
\bibfield{author}{\bibinfo{person}{OpenAI}, \bibinfo{person}{:}, \bibinfo{person}{Josh Achiam}, \bibinfo{person}{Steven Adler}, \bibinfo{person}{Sandhini Agarwal}, \bibinfo{person}{Lama Ahmad}, \bibinfo{person}{Ilge Akkaya}, \bibinfo{person}{Florencia~Leoni Aleman}, \bibinfo{person}{Diogo Almeida}, \bibinfo{person}{Janko Altenschmidt}, \bibinfo{person}{Sam Altman}, \bibinfo{person}{Shyamal Anadkat}, \bibinfo{person}{Red Avila}, \bibinfo{person}{Igor Babuschkin}, \bibinfo{person}{Suchir Balaji}, \bibinfo{person}{Valerie Balcom}, \bibinfo{person}{Paul Baltescu}, \bibinfo{person}{Haiming Bao}, \bibinfo{person}{Mo Bavarian}, \bibinfo{person}{Jeff Belgum}, \bibinfo{person}{Irwan Bello}, \bibinfo{person}{Jake Berdine}, \bibinfo{person}{Gabriel Bernadett-Shapiro}, \bibinfo{person}{Christopher Berner}, \bibinfo{person}{Lenny Bogdonoff}, \bibinfo{person}{Oleg Boiko}, \bibinfo{person}{Madelaine Boyd}, \bibinfo{person}{Anna-Luisa Brakman}, \bibinfo{person}{Greg Brockman}, \bibinfo{person}{Tim Brooks},
  \bibinfo{person}{Miles Brundage}, \bibinfo{person}{Kevin Button}, \bibinfo{person}{Trevor Cai}, \bibinfo{person}{Rosie Campbell}, \bibinfo{person}{Andrew Cann}, \bibinfo{person}{Brittany Carey}, \bibinfo{person}{Chelsea Carlson}, \bibinfo{person}{Rory Carmichael}, \bibinfo{person}{Brooke Chan}, \bibinfo{person}{Che Chang}, \bibinfo{person}{Fotis Chantzis}, \bibinfo{person}{Derek Chen}, \bibinfo{person}{Sully Chen}, \bibinfo{person}{Ruby Chen}, \bibinfo{person}{Jason Chen}, \bibinfo{person}{Mark Chen}, \bibinfo{person}{Ben Chess}, \bibinfo{person}{Chester Cho}, \bibinfo{person}{Casey Chu}, \bibinfo{person}{Hyung~Won Chung}, \bibinfo{person}{Dave Cummings}, \bibinfo{person}{Jeremiah Currier}, \bibinfo{person}{Yunxing Dai}, \bibinfo{person}{Cory Decareaux}, \bibinfo{person}{Thomas Degry}, \bibinfo{person}{Noah Deutsch}, \bibinfo{person}{Damien Deville}, \bibinfo{person}{Arka Dhar}, \bibinfo{person}{David Dohan}, \bibinfo{person}{Steve Dowling}, \bibinfo{person}{Sheila Dunning}, \bibinfo{person}{Adrien
  Ecoffet}, \bibinfo{person}{Atty Eleti}, \bibinfo{person}{Tyna Eloundou}, \bibinfo{person}{David Farhi}, \bibinfo{person}{Liam Fedus}, \bibinfo{person}{Niko Felix}, \bibinfo{person}{Simón~Posada Fishman}, \bibinfo{person}{Juston Forte}, \bibinfo{person}{Isabella Fulford}, \bibinfo{person}{Leo Gao}, \bibinfo{person}{Elie Georges}, \bibinfo{person}{Christian Gibson}, \bibinfo{person}{Vik Goel}, \bibinfo{person}{Tarun Gogineni}, \bibinfo{person}{Gabriel Goh}, \bibinfo{person}{Rapha Gontijo-Lopes}, \bibinfo{person}{Jonathan Gordon}, \bibinfo{person}{Morgan Grafstein}, \bibinfo{person}{Scott Gray}, \bibinfo{person}{Ryan Greene}, \bibinfo{person}{Joshua Gross}, \bibinfo{person}{Shixiang~Shane Gu}, \bibinfo{person}{Yufei Guo}, \bibinfo{person}{Chris Hallacy}, \bibinfo{person}{Jesse Han}, \bibinfo{person}{Jeff Harris}, \bibinfo{person}{Yuchen He}, \bibinfo{person}{Mike Heaton}, \bibinfo{person}{Johannes Heidecke}, \bibinfo{person}{Chris Hesse}, \bibinfo{person}{Alan Hickey}, \bibinfo{person}{Wade Hickey},
  \bibinfo{person}{Peter Hoeschele}, \bibinfo{person}{Brandon Houghton}, \bibinfo{person}{Kenny Hsu}, \bibinfo{person}{Shengli Hu}, \bibinfo{person}{Xin Hu}, \bibinfo{person}{Joost Huizinga}, \bibinfo{person}{Shantanu Jain}, \bibinfo{person}{Shawn Jain}, \bibinfo{person}{Joanne Jang}, \bibinfo{person}{Angela Jiang}, \bibinfo{person}{Roger Jiang}, \bibinfo{person}{Haozhun Jin}, \bibinfo{person}{Denny Jin}, \bibinfo{person}{Shino Jomoto}, \bibinfo{person}{Billie Jonn}, \bibinfo{person}{Heewoo Jun}, \bibinfo{person}{Tomer Kaftan}, \bibinfo{person}{Łukasz Kaiser}, \bibinfo{person}{Ali Kamali}, \bibinfo{person}{Ingmar Kanitscheider}, \bibinfo{person}{Nitish~Shirish Keskar}, \bibinfo{person}{Tabarak Khan}, \bibinfo{person}{Logan Kilpatrick}, \bibinfo{person}{Jong~Wook Kim}, \bibinfo{person}{Christina Kim}, \bibinfo{person}{Yongjik Kim}, \bibinfo{person}{Hendrik Kirchner}, \bibinfo{person}{Jamie Kiros}, \bibinfo{person}{Matt Knight}, \bibinfo{person}{Daniel Kokotajlo}, \bibinfo{person}{Łukasz Kondraciuk},
  \bibinfo{person}{Andrew Kondrich}, \bibinfo{person}{Aris Konstantinidis}, \bibinfo{person}{Kyle Kosic}, \bibinfo{person}{Gretchen Krueger}, \bibinfo{person}{Vishal Kuo}, \bibinfo{person}{Michael Lampe}, \bibinfo{person}{Ikai Lan}, \bibinfo{person}{Teddy Lee}, \bibinfo{person}{Jan Leike}, \bibinfo{person}{Jade Leung}, \bibinfo{person}{Daniel Levy}, \bibinfo{person}{Chak~Ming Li}, \bibinfo{person}{Rachel Lim}, \bibinfo{person}{Molly Lin}, \bibinfo{person}{Stephanie Lin}, \bibinfo{person}{Mateusz Litwin}, \bibinfo{person}{Theresa Lopez}, \bibinfo{person}{Ryan Lowe}, \bibinfo{person}{Patricia Lue}, \bibinfo{person}{Anna Makanju}, \bibinfo{person}{Kim Malfacini}, \bibinfo{person}{Sam Manning}, \bibinfo{person}{Todor Markov}, \bibinfo{person}{Yaniv Markovski}, \bibinfo{person}{Bianca Martin}, \bibinfo{person}{Katie Mayer}, \bibinfo{person}{Andrew Mayne}, \bibinfo{person}{Bob McGrew}, \bibinfo{person}{Scott~Mayer McKinney}, \bibinfo{person}{Christine McLeavey}, \bibinfo{person}{Paul McMillan},
  \bibinfo{person}{Jake McNeil}, \bibinfo{person}{David Medina}, \bibinfo{person}{Aalok Mehta}, \bibinfo{person}{Jacob Menick}, \bibinfo{person}{Luke Metz}, \bibinfo{person}{Andrey Mishchenko}, \bibinfo{person}{Pamela Mishkin}, \bibinfo{person}{Vinnie Monaco}, \bibinfo{person}{Evan Morikawa}, \bibinfo{person}{Daniel Mossing}, \bibinfo{person}{Tong Mu}, \bibinfo{person}{Mira Murati}, \bibinfo{person}{Oleg Murk}, \bibinfo{person}{David Mély}, \bibinfo{person}{Ashvin Nair}, \bibinfo{person}{Reiichiro Nakano}, \bibinfo{person}{Rajeev Nayak}, \bibinfo{person}{Arvind Neelakantan}, \bibinfo{person}{Richard Ngo}, \bibinfo{person}{Hyeonwoo Noh}, \bibinfo{person}{Long Ouyang}, \bibinfo{person}{Cullen O'Keefe}, \bibinfo{person}{Jakub Pachocki}, \bibinfo{person}{Alex Paino}, \bibinfo{person}{Joe Palermo}, \bibinfo{person}{Ashley Pantuliano}, \bibinfo{person}{Giambattista Parascandolo}, \bibinfo{person}{Joel Parish}, \bibinfo{person}{Emy Parparita}, \bibinfo{person}{Alex Passos}, \bibinfo{person}{Mikhail Pavlov},
  \bibinfo{person}{Andrew Peng}, \bibinfo{person}{Adam Perelman}, \bibinfo{person}{Filipe de Avila Belbute~Peres}, \bibinfo{person}{Michael Petrov}, \bibinfo{person}{Henrique~Ponde de Oliveira~Pinto}, \bibinfo{person}{Michael}, \bibinfo{person}{Pokorny}, \bibinfo{person}{Michelle Pokrass}, \bibinfo{person}{Vitchyr Pong}, \bibinfo{person}{Tolly Powell}, \bibinfo{person}{Alethea Power}, \bibinfo{person}{Boris Power}, \bibinfo{person}{Elizabeth Proehl}, \bibinfo{person}{Raul Puri}, \bibinfo{person}{Alec Radford}, \bibinfo{person}{Jack Rae}, \bibinfo{person}{Aditya Ramesh}, \bibinfo{person}{Cameron Raymond}, \bibinfo{person}{Francis Real}, \bibinfo{person}{Kendra Rimbach}, \bibinfo{person}{Carl Ross}, \bibinfo{person}{Bob Rotsted}, \bibinfo{person}{Henri Roussez}, \bibinfo{person}{Nick Ryder}, \bibinfo{person}{Mario Saltarelli}, \bibinfo{person}{Ted Sanders}, \bibinfo{person}{Shibani Santurkar}, \bibinfo{person}{Girish Sastry}, \bibinfo{person}{Heather Schmidt}, \bibinfo{person}{David Schnurr},
  \bibinfo{person}{John Schulman}, \bibinfo{person}{Daniel Selsam}, \bibinfo{person}{Kyla Sheppard}, \bibinfo{person}{Toki Sherbakov}, \bibinfo{person}{Jessica Shieh}, \bibinfo{person}{Sarah Shoker}, \bibinfo{person}{Pranav Shyam}, \bibinfo{person}{Szymon Sidor}, \bibinfo{person}{Eric Sigler}, \bibinfo{person}{Maddie Simens}, \bibinfo{person}{Jordan Sitkin}, \bibinfo{person}{Katarina Slama}, \bibinfo{person}{Ian Sohl}, \bibinfo{person}{Benjamin Sokolowsky}, \bibinfo{person}{Yang Song}, \bibinfo{person}{Natalie Staudacher}, \bibinfo{person}{Felipe~Petroski Such}, \bibinfo{person}{Natalie Summers}, \bibinfo{person}{Ilya Sutskever}, \bibinfo{person}{Jie Tang}, \bibinfo{person}{Nikolas Tezak}, \bibinfo{person}{Madeleine Thompson}, \bibinfo{person}{Phil Tillet}, \bibinfo{person}{Amin Tootoonchian}, \bibinfo{person}{Elizabeth Tseng}, \bibinfo{person}{Preston Tuggle}, \bibinfo{person}{Nick Turley}, \bibinfo{person}{Jerry Tworek}, \bibinfo{person}{Juan Felipe~Cerón Uribe}, \bibinfo{person}{Andrea Vallone},
  \bibinfo{person}{Arun Vijayvergiya}, \bibinfo{person}{Chelsea Voss}, \bibinfo{person}{Carroll Wainwright}, \bibinfo{person}{Justin~Jay Wang}, \bibinfo{person}{Alvin Wang}, \bibinfo{person}{Ben Wang}, \bibinfo{person}{Jonathan Ward}, \bibinfo{person}{Jason Wei}, \bibinfo{person}{CJ Weinmann}, \bibinfo{person}{Akila Welihinda}, \bibinfo{person}{Peter Welinder}, \bibinfo{person}{Jiayi Weng}, \bibinfo{person}{Lilian Weng}, \bibinfo{person}{Matt Wiethoff}, \bibinfo{person}{Dave Willner}, \bibinfo{person}{Clemens Winter}, \bibinfo{person}{Samuel Wolrich}, \bibinfo{person}{Hannah Wong}, \bibinfo{person}{Lauren Workman}, \bibinfo{person}{Sherwin Wu}, \bibinfo{person}{Jeff Wu}, \bibinfo{person}{Michael Wu}, \bibinfo{person}{Kai Xiao}, \bibinfo{person}{Tao Xu}, \bibinfo{person}{Sarah Yoo}, \bibinfo{person}{Kevin Yu}, \bibinfo{person}{Qiming Yuan}, \bibinfo{person}{Wojciech Zaremba}, \bibinfo{person}{Rowan Zellers}, \bibinfo{person}{Chong Zhang}, \bibinfo{person}{Marvin Zhang}, \bibinfo{person}{Shengjia Zhao},
  \bibinfo{person}{Tianhao Zheng}, \bibinfo{person}{Juntang Zhuang}, \bibinfo{person}{William Zhuk}, {and} \bibinfo{person}{Barret Zoph}.} \bibinfo{year}{2023}\natexlab{}.
\newblock \bibinfo{title}{GPT-4 Technical Report}.
\newblock
\newblock
\showeprint[arxiv]{2303.08774}~[cs.CL]


\bibitem[OpenAI(2022)]%
        {openai2022chatgpt}
\bibfield{author}{\bibinfo{person}{OpenAI}.} \bibinfo{year}{2022}\natexlab{}.
\newblock \showarticletitle{Introducing ChatGPT}.
\newblock \bibinfo{journal}{\emph{OpenAI Blog}} \bibinfo{volume}{{}}, \bibinfo{number}{{}} (\bibinfo{date}{Nov} \bibinfo{year}{2022}), \bibinfo{pages}{{}}.
\newblock


\bibitem[OpenAI(2024a)]%
        {openaimodels}
\bibfield{author}{\bibinfo{person}{OpenAI}.} \bibinfo{year}{2024}\natexlab{a}.
\newblock \bibinfo{title}{Models}.
\newblock \bibinfo{howpublished}{\url{https://platform.openai.com/docs/models/}}.
\newblock
\newblock
\shownote{[Accessed 19-01-2024]}.


\bibitem[OpenAI(2024b)]%
        {openaipricing}
\bibfield{author}{\bibinfo{person}{OpenAI}.} \bibinfo{year}{2024}\natexlab{b}.
\newblock \bibinfo{title}{Pricing}.
\newblock \bibinfo{howpublished}{\url{https://openai.com/pricing}}.
\newblock
\newblock
\shownote{[Accessed 19-01-2024]}.


\bibitem[Ouyang et~al\mbox{.}(2022)]%
        {ouyang2022training}
\bibfield{author}{\bibinfo{person}{Long Ouyang}, \bibinfo{person}{Jeffrey Wu}, \bibinfo{person}{Xu Jiang}, \bibinfo{person}{Diogo Almeida}, \bibinfo{person}{Carroll Wainwright}, \bibinfo{person}{Pamela Mishkin}, \bibinfo{person}{Chong Zhang}, \bibinfo{person}{Sandhini Agarwal}, \bibinfo{person}{Katarina Slama}, \bibinfo{person}{Alex Ray}, {et~al\mbox{.}}} \bibinfo{year}{2022}\natexlab{}.
\newblock \showarticletitle{Training language models to follow instructions with human feedback}.
\newblock \bibinfo{journal}{\emph{Advances in Neural Information Processing Systems}}  \bibinfo{volume}{35} (\bibinfo{year}{2022}), \bibinfo{pages}{27730--27744}.
\newblock


\bibitem[Palmer et~al\mbox{.}(2023)]%
        {palmer2023using}
\bibfield{author}{\bibinfo{person}{Alexis Palmer}, \bibinfo{person}{Noah~A Smith}, {and} \bibinfo{person}{Arthur Spirling}.} \bibinfo{year}{2023}\natexlab{}.
\newblock \showarticletitle{Using proprietary language models in academic research requires explicit justification}.
\newblock \bibinfo{journal}{\emph{Nature Computational Science}} \bibinfo{volume}{{}}, \bibinfo{number}{{}} (\bibinfo{year}{2023}), \bibinfo{pages}{1--2}.
\newblock


\bibitem[Papineni et~al\mbox{.}(2002)]%
        {papineni2002bleu}
\bibfield{author}{\bibinfo{person}{Kishore Papineni}, \bibinfo{person}{Salim Roukos}, \bibinfo{person}{Todd Ward}, {and} \bibinfo{person}{Wei-Jing Zhu}.} \bibinfo{year}{2002}\natexlab{}.
\newblock \showarticletitle{Bleu: a method for automatic evaluation of machine translation}. In \bibinfo{booktitle}{\emph{Proceedings of the 40th annual meeting of the Association for Computational Linguistics}}. \bibinfo{publisher}{{}}, \bibinfo{address}{{}}, \bibinfo{pages}{311--318}.
\newblock


\bibitem[Patel and Wong(2023)]%
        {gpt4architecture}
\bibfield{author}{\bibinfo{person}{Dylan Patel} {and} \bibinfo{person}{Gerald Wong}.} \bibinfo{year}{2023}\natexlab{}.
\newblock \bibinfo{title}{GPT-4 Architecture, Infrastructure, Training Dataset, Costs, Vision, MoE}.
\newblock \bibinfo{howpublished}{\url{https://www.semianalysis.com/p/gpt-4-architecture-infrastructure}}.
\newblock
\newblock
\shownote{[Accessed 19-01-2024]}.


\bibitem[Penedo et~al\mbox{.}(2023)]%
        {penedo2023refinedweb}
\bibfield{author}{\bibinfo{person}{Guilherme Penedo}, \bibinfo{person}{Quentin Malartic}, \bibinfo{person}{Daniel Hesslow}, \bibinfo{person}{Ruxandra Cojocaru}, \bibinfo{person}{Alessandro Cappelli}, \bibinfo{person}{Hamza Alobeidli}, \bibinfo{person}{Baptiste Pannier}, \bibinfo{person}{Ebtesam Almazrouei}, {and} \bibinfo{person}{Julien Launay}.} \bibinfo{year}{2023}\natexlab{}.
\newblock \showarticletitle{The RefinedWeb dataset for Falcon LLM: outperforming curated corpora with web data, and web data only}.
\newblock \bibinfo{journal}{\emph{arXiv preprint arXiv:2306.01116}} \bibinfo{volume}{{}}, \bibinfo{number}{{}} (\bibinfo{year}{2023}), \bibinfo{pages}{{}}.
\newblock


\bibitem[Peris et~al\mbox{.}(2023)]%
        {peris2023privacy}
\bibfield{author}{\bibinfo{person}{Charith Peris}, \bibinfo{person}{Christophe Dupuy}, \bibinfo{person}{Jimit Majmudar}, \bibinfo{person}{Rahil Parikh}, \bibinfo{person}{Sami Smaili}, \bibinfo{person}{Richard Zemel}, {and} \bibinfo{person}{Rahul Gupta}.} \bibinfo{year}{2023}\natexlab{}.
\newblock \showarticletitle{Privacy in the Time of Language Models}. In \bibinfo{booktitle}{\emph{Proceedings of the Sixteenth ACM International Conference on Web Search and Data Mining}}. \bibinfo{publisher}{{}}, \bibinfo{address}{{}}, \bibinfo{pages}{1291--1292}.
\newblock


\bibitem[Porter(2023)]%
        {porter2023chatgpt}
\bibfield{author}{\bibinfo{person}{Jon Porter}.} \bibinfo{year}{2023}\natexlab{}.
\newblock \showarticletitle{ChatGPT continues to be one of the fastest-growing services ever}.
\newblock \bibinfo{journal}{\emph{The Verge}} \bibinfo{volume}{{}}, \bibinfo{number}{{}} (\bibinfo{date}{Nov} \bibinfo{year}{2023}), \bibinfo{pages}{{}}.
\newblock


\bibitem[Radford et~al\mbox{.}(2018)]%
        {radford2018improving}
\bibfield{author}{\bibinfo{person}{Alec Radford}, \bibinfo{person}{Karthik Narasimhan}, \bibinfo{person}{Tim Salimans}, \bibinfo{person}{Ilya Sutskever}, {et~al\mbox{.}}} \bibinfo{year}{2018}\natexlab{}.
\newblock \showarticletitle{Improving language understanding by generative pre-training}.
\newblock \bibinfo{journal}{\emph{{}}} \bibinfo{volume}{{}}, \bibinfo{number}{{}} (\bibinfo{year}{2018}), \bibinfo{pages}{{}}.
\newblock


\bibitem[Radford et~al\mbox{.}(2019)]%
        {radford2019language}
\bibfield{author}{\bibinfo{person}{Alec Radford}, \bibinfo{person}{Jeffrey Wu}, \bibinfo{person}{Rewon Child}, \bibinfo{person}{David Luan}, \bibinfo{person}{Dario Amodei}, \bibinfo{person}{Ilya Sutskever}, {et~al\mbox{.}}} \bibinfo{year}{2019}\natexlab{}.
\newblock \showarticletitle{Language models are unsupervised multitask learners}.
\newblock \bibinfo{journal}{\emph{OpenAI blog}} \bibinfo{volume}{1}, \bibinfo{number}{8} (\bibinfo{year}{2019}), \bibinfo{pages}{9}.
\newblock


\bibitem[Rafailov et~al\mbox{.}(2023)]%
        {rafailov2023direct}
\bibfield{author}{\bibinfo{person}{Rafael Rafailov}, \bibinfo{person}{Archit Sharma}, \bibinfo{person}{Eric Mitchell}, \bibinfo{person}{Stefano Ermon}, \bibinfo{person}{Christopher~D Manning}, {and} \bibinfo{person}{Chelsea Finn}.} \bibinfo{year}{2023}\natexlab{}.
\newblock \showarticletitle{Direct preference optimization: Your language model is secretly a reward model}.
\newblock \bibinfo{journal}{\emph{arXiv preprint arXiv:2305.18290}} \bibinfo{volume}{{}}, \bibinfo{number}{{}} (\bibinfo{year}{2023}), \bibinfo{pages}{{}}.
\newblock


\bibitem[Ray(2023)]%
        {ray2023chatgpt}
\bibfield{author}{\bibinfo{person}{Partha~Pratim Ray}.} \bibinfo{year}{2023}\natexlab{}.
\newblock \showarticletitle{ChatGPT: A comprehensive review on background, applications, key challenges, bias, ethics, limitations and future scope}.
\newblock \bibinfo{journal}{\emph{Internet of Things and Cyber-Physical Systems}} \bibinfo{volume}{{}}, \bibinfo{number}{{}} (\bibinfo{year}{2023}), \bibinfo{pages}{{}}.
\newblock


\bibitem[Rogers et~al\mbox{.}(2023)]%
        {rogers-etal-2023-closed}
\bibfield{author}{\bibinfo{person}{Anna Rogers}, \bibinfo{person}{Niranjan Balasubramanian}, \bibinfo{person}{Leon Derczynski}, \bibinfo{person}{Jesse Dodge}, \bibinfo{person}{Alexander Koller}, \bibinfo{person}{Sasha Luccioni}, \bibinfo{person}{Maarten Sap}, \bibinfo{person}{Roy Schwartz}, \bibinfo{person}{Noah~A. Smith}, {and} \bibinfo{person}{Emma Strubell}.} \bibinfo{year}{2023}\natexlab{}.
\newblock \bibinfo{title}{Closed AI Models Make Bad Baselines}.
\newblock
\newblock
\urldef\tempurl%
\url{https://hackingsemantics.xyz/2023/closed-baselines/}
\showURL{%
\tempurl}


\bibitem[Sanh et~al\mbox{.}(2022)]%
        {sanh2022multitask}
\bibfield{author}{\bibinfo{person}{Victor Sanh}, \bibinfo{person}{Albert Webson}, \bibinfo{person}{Colin Raffel}, \bibinfo{person}{Stephen~H. Bach}, \bibinfo{person}{Lintang Sutawika}, \bibinfo{person}{Zaid Alyafeai}, \bibinfo{person}{Antoine Chaffin}, \bibinfo{person}{Arnaud Stiegler}, \bibinfo{person}{Teven Le~Scao}, \bibinfo{person}{Arun Raja}, \bibinfo{person}{Manan Dey}, \bibinfo{person}{M~Saiful Bari}, \bibinfo{person}{Canwen Xu}, \bibinfo{person}{Urmish Thakker}, \bibinfo{person}{Shanya Sharma}, \bibinfo{person}{Eliza Szczechla}, \bibinfo{person}{Taewoon Kim}, \bibinfo{person}{Gunjan Chhablani}, \bibinfo{person}{Nihal~V. Nayak}, \bibinfo{person}{Debajyoti Datta}, \bibinfo{person}{Jonathan Chang}, \bibinfo{person}{Mike Tian-Jian Jiang}, \bibinfo{person}{Han Wang}, \bibinfo{person}{Matteo Manica}, \bibinfo{person}{Sheng Shen}, \bibinfo{person}{Zheng-Xin Yong}, \bibinfo{person}{Harshit Pandey}, \bibinfo{person}{Michael McKenna}, \bibinfo{person}{Rachel Bawden}, \bibinfo{person}{Thomas Wang},
  \bibinfo{person}{Trishala Neeraj}, \bibinfo{person}{Jos Rozen}, \bibinfo{person}{Abheesht Sharma}, \bibinfo{person}{Andrea Santilli}, \bibinfo{person}{Thibault Fevry}, \bibinfo{person}{Jason~Alan Fries}, \bibinfo{person}{Ryan Teehan}, \bibinfo{person}{Tali Bers}, \bibinfo{person}{Stella Biderman}, \bibinfo{person}{Leo Gao}, \bibinfo{person}{Thomas Wolf}, {and} \bibinfo{person}{Alexander~M. Rush}.} \bibinfo{year}{2022}\natexlab{}.
\newblock \showarticletitle{Multitask Prompted Training Enables Zero-Shot Task Generalization}. In \bibinfo{booktitle}{\emph{International Conference on Learning Representations}}. \bibinfo{publisher}{{}}, \bibinfo{address}{{}}, \bibinfo{pages}{{}}.
\newblock
\urldef\tempurl%
\url{https://research.ibm.com/publications/multitask-prompt-tuning-enables-zero-shot-task-generalization}
\showURL{%
\tempurl}


\bibitem[Thalken et~al\mbox{.}(2023)]%
        {thalken2023modeling}
\bibfield{author}{\bibinfo{person}{Rosamond Thalken}, \bibinfo{person}{Edward Stiglitz}, \bibinfo{person}{David Mimno}, {and} \bibinfo{person}{Matthew Wilkens}.} \bibinfo{year}{2023}\natexlab{}.
\newblock \showarticletitle{Modeling Legal Reasoning: LM Annotation at the Edge of Human Agreement}. In \bibinfo{booktitle}{\emph{Proceedings of the 2023 Conference on Empirical Methods in Natural Language Processing}}. \bibinfo{publisher}{{}}, \bibinfo{address}{{}}, \bibinfo{pages}{9252--9265}.
\newblock


\bibitem[Touvron et~al\mbox{.}(2023a)]%
        {touvron2023llama}
\bibfield{author}{\bibinfo{person}{Hugo Touvron}, \bibinfo{person}{Thibaut Lavril}, \bibinfo{person}{Gautier Izacard}, \bibinfo{person}{Xavier Martinet}, \bibinfo{person}{Marie-Anne Lachaux}, \bibinfo{person}{Timoth{\'e}e Lacroix}, \bibinfo{person}{Baptiste Rozi{\`e}re}, \bibinfo{person}{Naman Goyal}, \bibinfo{person}{Eric Hambro}, \bibinfo{person}{Faisal Azhar}, {et~al\mbox{.}}} \bibinfo{year}{2023}\natexlab{a}.
\newblock \showarticletitle{Llama: Open and efficient foundation language models}.
\newblock \bibinfo{journal}{\emph{arXiv preprint arXiv:2302.13971}} \bibinfo{volume}{{}}, \bibinfo{number}{{}} (\bibinfo{year}{2023}), \bibinfo{pages}{{}}.
\newblock


\bibitem[Touvron et~al\mbox{.}(2023b)]%
        {touvron2023llama2}
\bibfield{author}{\bibinfo{person}{Hugo Touvron}, \bibinfo{person}{Louis Martin}, \bibinfo{person}{Kevin Stone}, \bibinfo{person}{Peter Albert}, \bibinfo{person}{Amjad Almahairi}, \bibinfo{person}{Yasmine Babaei}, \bibinfo{person}{Nikolay Bashlykov}, \bibinfo{person}{Soumya Batra}, \bibinfo{person}{Prajjwal Bhargava}, \bibinfo{person}{Shruti Bhosale}, {et~al\mbox{.}}} \bibinfo{year}{2023}\natexlab{b}.
\newblock \showarticletitle{Llama 2: Open foundation and fine-tuned chat models}.
\newblock \bibinfo{journal}{\emph{arXiv preprint arXiv:2307.09288}} \bibinfo{volume}{{}}, \bibinfo{number}{{}} (\bibinfo{year}{2023}), \bibinfo{pages}{{}}.
\newblock


\bibitem[Tu et~al\mbox{.}(2023)]%
        {tu2023towards}
\bibfield{author}{\bibinfo{person}{Tao Tu}, \bibinfo{person}{Shekoofeh Azizi}, \bibinfo{person}{Danny Driess}, \bibinfo{person}{Mike Schaekermann}, \bibinfo{person}{Mohamed Amin}, \bibinfo{person}{Pi-Chuan Chang}, \bibinfo{person}{Andrew Carroll}, \bibinfo{person}{Chuck Lau}, \bibinfo{person}{Ryutaro Tanno}, \bibinfo{person}{Ira Ktena}, {et~al\mbox{.}}} \bibinfo{year}{2023}\natexlab{}.
\newblock \showarticletitle{Towards generalist biomedical AI}.
\newblock \bibinfo{journal}{\emph{arXiv preprint arXiv:2307.14334}} \bibinfo{volume}{{}}, \bibinfo{number}{{}} (\bibinfo{year}{2023}), \bibinfo{pages}{{}}.
\newblock


\bibitem[Vaswani et~al\mbox{.}(2017)]%
        {vaswani2017attention}
\bibfield{author}{\bibinfo{person}{Ashish Vaswani}, \bibinfo{person}{Noam Shazeer}, \bibinfo{person}{Niki Parmar}, \bibinfo{person}{Jakob Uszkoreit}, \bibinfo{person}{Llion Jones}, \bibinfo{person}{Aidan~N Gomez}, \bibinfo{person}{{\L}ukasz Kaiser}, {and} \bibinfo{person}{Illia Polosukhin}.} \bibinfo{year}{2017}\natexlab{}.
\newblock \showarticletitle{Attention is all you need}.
\newblock \bibinfo{journal}{\emph{Advances in neural information processing systems}} \bibinfo{volume}{30}, \bibinfo{number}{{}} (\bibinfo{year}{2017}), \bibinfo{pages}{{}}.
\newblock


\bibitem[Waisberg et~al\mbox{.}(2023)]%
        {waisberg2023gpt}
\bibfield{author}{\bibinfo{person}{Ethan Waisberg}, \bibinfo{person}{Joshua Ong}, \bibinfo{person}{Nasif Zaman}, \bibinfo{person}{Sharif~Amit Kamran}, \bibinfo{person}{Prithul Sarker}, \bibinfo{person}{Alireza Tavakkoli}, {and} \bibinfo{person}{Andrew~G Lee}.} \bibinfo{year}{2023}\natexlab{}.
\newblock \showarticletitle{GPT-4 for triaging ophthalmic symptoms}.
\newblock \bibinfo{journal}{\emph{Eye}} \bibinfo{volume}{37}, \bibinfo{number}{18} (\bibinfo{year}{2023}), \bibinfo{pages}{3874--3875}.
\newblock


\bibitem[Wang et~al\mbox{.}(2019)]%
        {wang2019survey}
\bibfield{author}{\bibinfo{person}{Wei Wang}, \bibinfo{person}{Vincent~W Zheng}, \bibinfo{person}{Han Yu}, {and} \bibinfo{person}{Chunyan Miao}.} \bibinfo{year}{2019}\natexlab{}.
\newblock \showarticletitle{A survey of zero-shot learning: Settings, methods, and applications}.
\newblock \bibinfo{journal}{\emph{ACM Transactions on Intelligent Systems and Technology (TIST)}} \bibinfo{volume}{10}, \bibinfo{number}{2} (\bibinfo{year}{2019}), \bibinfo{pages}{1--37}.
\newblock


\bibitem[Wang et~al\mbox{.}(2023)]%
        {wang2023chatgpt}
\bibfield{author}{\bibinfo{person}{Zengzhi Wang}, \bibinfo{person}{Qiming Xie}, \bibinfo{person}{Zixiang Ding}, \bibinfo{person}{Yi Feng}, {and} \bibinfo{person}{Rui Xia}.} \bibinfo{year}{2023}\natexlab{}.
\newblock \showarticletitle{Is ChatGPT a good sentiment analyzer? A preliminary study}.
\newblock \bibinfo{journal}{\emph{arXiv preprint arXiv:2304.04339}} \bibinfo{volume}{{}}, \bibinfo{number}{{}} (\bibinfo{year}{2023}), \bibinfo{pages}{{}}.
\newblock


\bibitem[Webersinke et~al\mbox{.}(2021)]%
        {webersinke2021climatebert}
\bibfield{author}{\bibinfo{person}{Nicolas Webersinke}, \bibinfo{person}{Mathias Kraus}, \bibinfo{person}{Julia~Anna Bingler}, {and} \bibinfo{person}{Markus Leippold}.} \bibinfo{year}{2021}\natexlab{}.
\newblock \showarticletitle{Climatebert: A pretrained language model for climate-related text}.
\newblock \bibinfo{journal}{\emph{arXiv preprint arXiv:2110.12010}} \bibinfo{volume}{{}}, \bibinfo{number}{{}} (\bibinfo{year}{2021}), \bibinfo{pages}{{}}.
\newblock


\bibitem[Wei et~al\mbox{.}(2022)]%
        {wei2022chain}
\bibfield{author}{\bibinfo{person}{Jason Wei}, \bibinfo{person}{Xuezhi Wang}, \bibinfo{person}{Dale Schuurmans}, \bibinfo{person}{Maarten Bosma}, \bibinfo{person}{Fei Xia}, \bibinfo{person}{Ed Chi}, \bibinfo{person}{Quoc~V Le}, \bibinfo{person}{Denny Zhou}, {et~al\mbox{.}}} \bibinfo{year}{2022}\natexlab{}.
\newblock \showarticletitle{Chain-of-thought prompting elicits reasoning in large language models}.
\newblock \bibinfo{journal}{\emph{Advances in Neural Information Processing Systems}}  \bibinfo{volume}{35} (\bibinfo{year}{2022}), \bibinfo{pages}{24824--24837}.
\newblock


\bibitem[Wen et~al\mbox{.}(2024)]%
        {wen2024characterizing}
\bibfield{author}{\bibinfo{person}{Bingbing Wen}, \bibinfo{person}{Bill Howe}, {and} \bibinfo{person}{Lucy~Lu Wang}.} \bibinfo{year}{2024}\natexlab{}.
\newblock \showarticletitle{Characterizing LLM Abstention Behavior in Science QA with Context Perturbations}.
\newblock \bibinfo{journal}{\emph{arXiv preprint arXiv:2404.12452}} (\bibinfo{year}{2024}).
\newblock


\bibitem[Wolf et~al\mbox{.}(2019)]%
        {wolf2019huggingface}
\bibfield{author}{\bibinfo{person}{Thomas Wolf}, \bibinfo{person}{Lysandre Debut}, \bibinfo{person}{Victor Sanh}, \bibinfo{person}{Julien Chaumond}, \bibinfo{person}{Clement Delangue}, \bibinfo{person}{Anthony Moi}, \bibinfo{person}{Pierric Cistac}, \bibinfo{person}{Tim Rault}, \bibinfo{person}{R{\'e}mi Louf}, \bibinfo{person}{Morgan Funtowicz}, {et~al\mbox{.}}} \bibinfo{year}{2019}\natexlab{}.
\newblock \showarticletitle{Huggingface's transformers: State-of-the-art natural language processing}.
\newblock \bibinfo{journal}{\emph{arXiv preprint arXiv:1910.03771}} \bibinfo{volume}{{}}, \bibinfo{number}{{}} (\bibinfo{year}{2019}), \bibinfo{pages}{{}}.
\newblock


\bibitem[Wutschitz et~al\mbox{.}(2022)]%
        {dp-transformers}
\bibfield{author}{\bibinfo{person}{Lukas Wutschitz}, \bibinfo{person}{Huseyin~A. Inan}, {and} \bibinfo{person}{Andre Manoel}.} \bibinfo{year}{2022}\natexlab{}.
\newblock \bibinfo{title}{dp-transformers: Training transformer models with differential privacy}.
\newblock \bibinfo{howpublished}{\url{https://www.microsoft.com/en-us/research/project/dp-transformers}}.
\newblock


\bibitem[Yim et~al\mbox{.}(2023)]%
        {yim2023overview}
\bibfield{author}{\bibinfo{person}{Wen-wai Yim}, \bibinfo{person}{A~Ben Abacha}, \bibinfo{person}{N Snider}, \bibinfo{person}{G Adams}, {and} \bibinfo{person}{Meliha Yetisgen}.} \bibinfo{year}{2023}\natexlab{}.
\newblock \showarticletitle{Overview of the mediqa-sum task at imageclef 2023: Summarization and classification of doctor-patient conversations}. In \bibinfo{booktitle}{\emph{CLEF}}. \bibinfo{publisher}{{}}, \bibinfo{address}{{}}, \bibinfo{pages}{{}}.
\newblock


\bibitem[Yousefpour et~al\mbox{.}(2021)]%
        {yousefpour2021opacus}
\bibfield{author}{\bibinfo{person}{Ashkan Yousefpour}, \bibinfo{person}{Igor Shilov}, \bibinfo{person}{Alexandre Sablayrolles}, \bibinfo{person}{Davide Testuggine}, \bibinfo{person}{Karthik Prasad}, \bibinfo{person}{Mani Malek}, \bibinfo{person}{John Nguyen}, \bibinfo{person}{Sayan Ghosh}, \bibinfo{person}{Akash Bharadwaj}, \bibinfo{person}{Jessica Zhao}, {et~al\mbox{.}}} \bibinfo{year}{2021}\natexlab{}.
\newblock \showarticletitle{Opacus: User-friendly differential privacy library in PyTorch}.
\newblock \bibinfo{journal}{\emph{arXiv preprint arXiv:2109.12298}} \bibinfo{volume}{{}}, \bibinfo{number}{{}} (\bibinfo{year}{2021}), \bibinfo{pages}{{}}.
\newblock


\bibitem[Yu et~al\mbox{.}(2021)]%
        {yu2021differentially}
\bibfield{author}{\bibinfo{person}{Da Yu}, \bibinfo{person}{Saurabh Naik}, \bibinfo{person}{Arturs Backurs}, \bibinfo{person}{Sivakanth Gopi}, \bibinfo{person}{Huseyin~A Inan}, \bibinfo{person}{Gautam Kamath}, \bibinfo{person}{Janardhan Kulkarni}, \bibinfo{person}{Yin~Tat Lee}, \bibinfo{person}{Andre Manoel}, \bibinfo{person}{Lukas Wutschitz}, {et~al\mbox{.}}} \bibinfo{year}{2021}\natexlab{}.
\newblock \showarticletitle{Differentially private fine-tuning of language models}.
\newblock \bibinfo{journal}{\emph{arXiv preprint arXiv:2110.06500}} \bibinfo{volume}{{}}, \bibinfo{number}{{}} (\bibinfo{year}{2021}), \bibinfo{pages}{{}}.
\newblock


\bibitem[Zaken et~al\mbox{.}(2022)]%
        {zaken2022bitfit}
\bibfield{author}{\bibinfo{person}{Elad~Ben Zaken}, \bibinfo{person}{Yoav Goldberg}, {and} \bibinfo{person}{Shauli Ravfogel}.} \bibinfo{year}{2022}\natexlab{}.
\newblock \showarticletitle{BitFit: Simple Parameter-efficient Fine-tuning for Transformer-based Masked Language-models}. In \bibinfo{booktitle}{\emph{Proceedings of the 60th Annual Meeting of the Association for Computational Linguistics (Volume 2: Short Papers)}}. \bibinfo{publisher}{{}}, \bibinfo{address}{{}}, \bibinfo{pages}{1--9}.
\newblock


\bibitem[Zapf et~al\mbox{.}(2016)]%
        {zapf_measuring_2016}
\bibfield{author}{\bibinfo{person}{Antonia Zapf}, \bibinfo{person}{Stefanie Castell}, \bibinfo{person}{Lars Morawietz}, {and} \bibinfo{person}{André Karch}.} \bibinfo{year}{2016}\natexlab{}.
\newblock \showarticletitle{Measuring inter-rater reliability for nominal data – which coefficients and confidence intervals are appropriate?}
\newblock \bibinfo{journal}{\emph{BMC Medical Research Methodology}} \bibinfo{volume}{16}, \bibinfo{number}{1} (\bibinfo{date}{Aug.} \bibinfo{year}{2016}), \bibinfo{pages}{93}.
\newblock
\showISSN{1471-2288}
\urldef\tempurl%
\url{https://doi.org/10.1186/s12874-016-0200-9}
\showDOI{\tempurl}


\bibitem[Zhang et~al\mbox{.}(2019)]%
        {zhang2019bertscore}
\bibfield{author}{\bibinfo{person}{Tianyi Zhang}, \bibinfo{person}{Varsha Kishore}, \bibinfo{person}{Felix Wu}, \bibinfo{person}{Kilian~Q Weinberger}, {and} \bibinfo{person}{Yoav Artzi}.} \bibinfo{year}{2019}\natexlab{}.
\newblock \showarticletitle{BERTScore: Evaluating Text Generation with BERT}. In \bibinfo{booktitle}{\emph{International Conference on Learning Representations}}. \bibinfo{publisher}{{}}, \bibinfo{address}{{}}, \bibinfo{pages}{{}}.
\newblock


\bibitem[Zheng et~al\mbox{.}(2023)]%
        {zheng2023judging}
\bibfield{author}{\bibinfo{person}{Lianmin Zheng}, \bibinfo{person}{Wei-Lin Chiang}, \bibinfo{person}{Ying Sheng}, \bibinfo{person}{Siyuan Zhuang}, \bibinfo{person}{Zhanghao Wu}, \bibinfo{person}{Yonghao Zhuang}, \bibinfo{person}{Zi Lin}, \bibinfo{person}{Zhuohan Li}, \bibinfo{person}{Dacheng Li}, \bibinfo{person}{Eric~P. Xing}, \bibinfo{person}{Hao Zhang}, \bibinfo{person}{Joseph~E. Gonzalez}, {and} \bibinfo{person}{Ion Stoica}.} \bibinfo{year}{2023}\natexlab{}.
\newblock \bibinfo{title}{Judging LLM-as-a-Judge with MT-Bench and Chatbot Arena}.
\newblock
\newblock
\showeprint[arxiv]{2306.05685}~[cs.CL]


\end{thebibliography}

\end{document}